%% file: main.tex
\documentclass{article}

\PassOptionsToPackage{numbers, compress}{natbib}

\usepackage[preprint]{neurips_2025} % None, preprint, final

\usepackage[utf8]{inputenc} % allow utf-8 input
\usepackage[T1]{fontenc}    % use 8-bit T1 fonts
\usepackage{hyperref}       % hyperlinks
\usepackage{url}            % simple URL typesetting
\usepackage{booktabs}       % professional-quality tables
\usepackage{amsfonts}       % blackboard math symbols
\usepackage{amsmath}
\usepackage{amssymb}
\usepackage{mathtools}
\usepackage{amsthm}
\usepackage{graphicx}
\usepackage{nicefrac}       % compact symbols for 1/2, etc.
\usepackage{microtype}      % microtypography
\usepackage{multicol,multirow}
\usepackage[table]{xcolor}
\usepackage{colortbl}
\usepackage{booktabs}
\usepackage{graphicx} % For \resizebox
\usepackage{bbm}
\usepackage{enumitem}
\usepackage{float}
\usepackage{subfig}
\usepackage{cleveref}

\crefformat{footnote}{#2\footnotemark[#1]#3}

\title{Gym4ReaL: A Suite for Benchmarking\\Real-World Reinforcement Learning}

\author{
  Davide Salaorni$^{1}$ 
  \quad
  Vincenzo De Paola$^{1}$ 
  \quad
  Samuele Delpero$^{1}$ 
  \quad
  Giovanni Dispoto$^{1}$ 
 \\
  \textbf{Paolo Bonetti}$^{1}$ 
  \quad
  \textbf{Alessio Russo}$^{1}$ 
  \quad
  \textbf{Giuseppe Calcagno}$^{1}$ 
  \quad
  \textbf{Francesco Trov\`{o}}$^{1}$ 
  \\
  \textbf{Matteo Papini}$^{1}$ 
  \quad
  \textbf{Alberto Maria Metelli}$^{1}$ 
  \quad
  \textbf{Marco Mussi}$^{1}$ 
  \quad
  \textbf{Marcello Restelli}$^{1}$ 
  \\
  $^{1}$ Politecnico di Milano \\
  mail to: \texttt{davide.salaorni@polimi.it}
}

\begin{document}

\allowdisplaybreaks[4]
\setlength{\abovedisplayskip}{4pt}
\setlength{\belowdisplayskip}{4pt}
\setlength{\textfloatsep}{10pt}

\maketitle

\begin{abstract}
In recent years, \emph{Reinforcement Learning} (RL) has made remarkable progress, achieving superhuman performance in a wide range of simulated environments. As research moves toward deploying RL in real-world applications, the field faces a new set of challenges inherent to real-world settings, such as large state-action spaces, non-stationarity, and partial observability. Despite their importance, these challenges are often underexplored in current benchmarks, which tend to focus on idealized, fully observable, and stationary environments, often neglecting to incorporate real-world complexities explicitly. In this paper, we introduce \texttt{Gym4ReaL}, a comprehensive suite of realistic environments designed to support the development and evaluation of RL algorithms that can operate in real-world scenarios. The suite includes a diverse set of tasks that expose algorithms to a variety of practical challenges. Our experimental results show that, in these settings, standard RL algorithms confirm their competitiveness against rule-based benchmarks, motivating the development of new methods to fully exploit the potential of RL to tackle the complexities of real-world tasks.

% , off-the-shelf

% Over the past decades, reinforcement learning (RL) has attracted growing attention, driven by the widespread integration of artificial intelligence into our daily life. However, the majority of benchmarking tasks used to evaluate novel RL algorithms are still based on toy problems or virtual environments. Indeed, a well-known limitation in RL research is that success in these artificial settings does not readily translate to effectiveness in real-world applications, affected by uncertainty, physical constraints, and human-world limitations. In this paper, we introduce Gym4ReaL, a comprehensive suite of realistic environments designed to support the development and evaluation of RL algorithms in real-world scenarios. The suite includes a diverse set of tasks that expose algorithms to a variety of practical challenges. Our experimental results show that standard, off-the-shelf RL algorithms still fall short in these settings, highlighting the need for new methods capable of tackling the complexities of real-world tasks.
\end{abstract}

\input{contents/introduction}
\input{contents/environments}

\input{contents/conclusions}

% Acknowledgments
%\input{contents/acks}

%\clearpage
\bibliography{refs}
\bibliographystyle{abbrvnat}

%%%%%%%%%%%%%%%%%%%%%%%%%%%%%%%%%%%%%%%%%%%%%%%%%%%%%%%%%%%%

\clearpage
\appendix
\input{appendix/metadata}
\input{appendix/deep_dive_table}
\input{appendix/datasets}
\input{appendix/env_details}
\input{appendix/experiment_details}

%%%%%%%%%%%%%%%%%%%%%%%%%%%%%%%%%%%%%%%%%%%%%%%%%%%%%%%%%%%%

%\clearpage
%\input{checklist}

\end{document}

%% file: contents/introduction.tex
\section{Introduction}
In the past few years, \emph{Reinforcement Learning} (RL)~\cite{Sutton1998} has demonstrated above-human performance across different challenges, ranging from playing Atari games~\cite{Mnih2015Atari} to beating world champions of Chess and Go~\cite{silver2017alphaZero, Silver2017alphaGo}, achieving impressive results also in the field of robotic control~\cite{kober2013reinforcement}. However, despite these promising advances, RL still struggles to gain traction in many real-world applications, where systems are often subject to uncertainties and unpredictable factors that complicate accurate physical modeling. An additional limitation lies in the fact that RL algorithms are typically validated on idealized environments, such as those provided by Gymnasium~\citep{towers2024gymnasium} and MuJoCo~\cite{todorov2012mujoco}.
%, which lack the inherent complexity of real-world tasks. 
Despite their great contribution to RL research, such libraries provide artificial playgrounds able to generate infinite samples, adapt to any desired configuration, and grant harmless exploration. However, learning and overfitting these environments does not necessarily reflect skillfulness in real-world tasks, where data is limited, dynamics change, and exploration does not come for free.
%As a result, the validation of new RL techniques remains incomplete, and their performance in practical settings often fails to match the results achieved in standard benchmarking suites.

As a step towards bridging the gap between simulated and real-world settings and promoting the deployment of RL in practical applications, we present \texttt{Gym4ReaL}\footnote{Codebase available here: \url{https://github.com/Daveonwave/gym4ReaL}}, a benchmarking suite designed to realistically model several real-world environments specifically tailored for RL algorithms. The selected tasks span multiple application domains. In particular, the suite includes:
\begin{itemize}[noitemsep, leftmargin=*, topsep=-1pt]
    \item \texttt{DamEnv}, which models a dam control system responsible for releasing the appropriate amount of water to meet residential demand;
    \item \texttt{ElevatorEnv}, which addresses a simplified version of the elevator dispatching problem under dynamic request patterns; 
    \item \texttt{MicrogridEnv}, which focuses on optimal energy management within a local microgrid, balancing supply, demand, and storage;
    \item \texttt{RoboFeederEnv}, which simulates a robotic work cell tasked with isolating and picking small objects, including both picking and planning challenges;
    \item \texttt{TradingEnv}, which aims to develop optimized trading strategies for the foreign exchange (Forex) market;
    \item \texttt{WDSEnv}, which models a municipal water distribution system, where the objective is to ensure a consistent supply to meet fluctuating residential demand.
\end{itemize}

While the tasks of these environments can be modeled in various ways, \texttt{Gym4ReaL} provides their standardized implementation compatible with  Gymnasium~\cite{towers2024gymnasium} interfaces. Our selection of real-world environments allows for training agents that specifically address such practical problems. However, \texttt{Gym4ReaL} is also designed to serve as a methodologically agnostic suite, enabling RL researchers to evaluate and benchmark algorithms regardless of any specific domain knowledge.

\input{contents/featuresTable}

The primary goal of \texttt{Gym4ReaL} is not merely to supply environments for solving specific domain tasks, but to offer a curated suite of environments that encapsulates the fundamental challenges inherent to these real-world environments. A comprehensive summary of the suite’s features is presented in Table~\ref{tab:benchmark_features}. In particular, we distinguish between two key aspects: \emph{characteristics}, which refer to modeling properties specific to each environment, and \emph{RL paradigms}, which denote the classes of RL techniques that can be effectively tested and benchmarked within these environments beyond the classical RL approaches.
Furthermore, \texttt{Gym4ReaL} offers a high degree of configurability. Users can customize input parameters and environmental dynamics to better reflect domain-specific requirements, thus extending the suite’s usability to researchers from the respective application domains. 
Through this combination of realism, diversity, and flexibility, \texttt{Gym4ReaL} supports a wide spectrum of research efforts, from benchmarking general-purpose RL algorithms under realistic conditions to developing domain-specific controllers.

%In addition, we also offer several customization options, allowing users to modify input parameters to better align the environments with domain-specific requirements, by making the suite accessible not only to RL researchers but also to specialists in the respective fields. Indeed, Gym4ReaL ensures a high-level of simulation accuracy, as most environments are built on high-fidelity simulators tailored to the specific task. For example, the \texttt{MicrogridEnv} leverages a battery digital twin to emulate the battery evolution over time~\cite{salaorni2023ernesto}, the \texttt{RoboFeederEnv} employs the Mujoco physics engine~\cite{depaola2024power}, and \texttt{WDSEnv} is based on Epanet~\cite{epanet2000}, a widely used hydraulic analysis toolkit for modeling water supply networks.

\textbf{Related works.}~~ 
Several prior works have introduced benchmarking suites aimed at evaluating RL algorithms in realistic or application-specific scenarios. 
For example, SustainGym~\cite{yeh2022sustaingym} provides a suite of energy system simulations focused on sustainability and grid optimization, designed to assess RL agents under real-world constraints such as energy storage dynamics and carbon emissions. Similarly, SustainDC~\cite{naug2024sustaidc} targets sustainable control in data centers, offering environments that capture the complexity of workload scheduling and energy-efficient operations. 
In the domain of autonomous driving, MetaDrive~\cite{li2021metadrive} offers a customizable simulator for training RL agents in diverse driving scenarios, testing generalization and robustness capabilities by supporting procedurally generated environments. Regarding financial problems, FinRL~\cite{liu2021finrl} proposes a framework that reduces the complexity of training RL agents to perform different financial tasks (e.g., trading and portfolio allocation) on different markets. 
In contrast to the aforementioned libraries, which focus on domain-specific tasks, \texttt{Gym4ReaL} is designed as a comprehensive benchmarking suite spanning multiple domains and aiming to serve as a reference point for RL research in real-world scenarios.

Beyond domain-specific suites, several other platforms aim to expose RL algorithms to broader real-world challenges. Real-World RL (RWRL) Suite \cite{dulacarnold2020realworldrlempirical} focuses on incorporating key aspects of real-world deployment such as partial observability, delayed rewards, and non-stationarity. Robust RL Suite  (RRLS) \cite{zouitine2024rrlsrobustreinforcement} introduces environmental uncertainty to evaluate the robustness of RL methods in continuous control tasks. POPGym~\cite{morad2023popgym} presents a set of environments centered on partially observable Markov decision processes (POMDPs), making it useful for studying memory and inference in RL.
Finally, Safe-Control-Gym~\cite{yuan2021safecontrolgym} provides environments tailored for the safe RL paradigm in robotics and control, incorporating constraints and safety-aware objectives.
Unlike \texttt{Gym4ReaL}, such suites do not aim to address real-world problems directly. Instead, they leverage virtual environments from Gymnasium and Mujoco to simulate some of the challenges of real-world tasks.

%% file: contents/featuresTable.tex
\renewcommand{\arraystretch}{1.1} % Adjust row spacing
\begin{table}[t]
    \centering
        \caption{\emph{Characteristics} and \emph{RL paradigms} covered by each environment provided by \texttt{Gym4ReaL}.}
    \rowcolors{2}{gray!10}{white} % Alternate row colors
    \begin{tabular}{l || *{6}{c} | *{6}{c}}
        \rowcolor{gray!30}
         & \multicolumn{6}{c}{\textbf{Characteristics}} 
         & \multicolumn{6}{|c}{\textbf{RL Paradigms}} \\
        \rowcolor{gray!15}
         & \rotatebox{90}{Cont. States}
         & \rotatebox{90}{Cont. Actions}   
         & \rotatebox{90}{Part. Observable} 
         & \rotatebox{90}{Part. Controllable} 
         & \rotatebox{90}{Non-Stationary} 
         & \rotatebox{90}{Visual Input} 
         & \rotatebox{90}{Freq. Adaptation} 
         & \rotatebox{90}{Hierarchical RL} 
         & \rotatebox{90}{Risk-Averse}
         & \rotatebox{90}{Imitation Learning}
         & \rotatebox{90}{Provably Efficient}
         & \rotatebox{90}{Multi-Objective RL}\\
         \toprule
         \texttt{DamEnv}  & \checkmark & \checkmark &  & \checkmark &  &  &  &  &  & \checkmark &  & \checkmark\\
         \texttt{ElevatorEnv}  &  &  &  & \checkmark &  &  &  &  &  &  & \checkmark &\\
         \texttt{MicrogridEnv}  & \checkmark & \checkmark  &  & \checkmark &  &  & \checkmark &  &  &  &  & \checkmark \\
         \texttt{RoboFeederEnv}  & \checkmark & \checkmark  &  &  &  & \checkmark &  & \checkmark &  &  &  &\\
         \texttt{TradingEnv}    & \checkmark  &  & \checkmark  & \checkmark & \checkmark &  & \checkmark  &  & \checkmark  &  &  &\\ 
         \texttt{WDSEnv}  & \checkmark &  &  & \checkmark &  &  &  &  &  & \checkmark &  & \checkmark\\
         \bottomrule
    \end{tabular}
    \label{tab:benchmark_features}
\end{table}

%% file: contents/environments.tex
\section{Environments and Benchmarking}
This section introduces \texttt{Gym4ReaL} environments, describing the state space, the action space, and the reward function. Test results associated with each task are included, comparing rule-based expert policies with RL-based agents. Further details on environments and results are in the Appendix.

\input{contents/envs/dam}
\input{contents/envs/elevator}
\input{contents/envs/microgrid}
\input{contents/envs/robofeeder}
\input{contents/envs/trading}
\input{contents/envs/wds}

%% file: contents/envs/dam.tex
\subsection{\texttt{DamEnv}} \label{sec:dam}
\texttt{DamEnv} is designed to model the operation of a dam connected to a water reservoir. By providing the amount of water to be released as an action, the environment simulates changes in the water level, considering inflows, outflows, and other physical dynamics. The agent controlling the dam aims to plan the water release in order to satisfy the daily water demand while preventing the reservoir from exceeding its maximum capacity and causing overflows.
%Formally, considering a time horizon of $\mathcal{T}$, the \emph{total objective reward} of the agent following policy $\pi = (a_0, \ldots, a_{\mathcal{T}})$ is:
%\begin{equation}
%    R_{\mathcal{T}}(\pi) = \sum_{t=1}^\mathcal{T} \ [r_{d}(a_t) + r_{of}(a_t) + r_{st}(a_t)] \label{eq:dam_objective},
%\end{equation}
Formally, the objective is: %the maximization of the following cumulative function:
% R_{\mathcal{T}}(\pi) = 
\begin{equation}
\max\sum_{t=1}^T \ [r_{d}(a_t) + r_{\text{of}}(a_t) + r_{\text{st}}(a_t)] \label{eq:dam_objective},
\end{equation}
where $r_{d}$ favors actions that meet daily demand, $r_{\text{of}}$ actions that prevent water overflows, and $r_{\text{st}}$ those that avoid starvation effects along the time horizon $T$. %Detailed definitions can be found in Appendix.
The daily control frequency adopted depends on the data granularity. Moreover, the available historical data derived from human-expert decisions allows for the development of imitation learning studies. 

%of the environment is daily and the horizon is%one day, although the internal simulation of the dam and reservoir can be set at initialization in a more fine-grained way.

\textbf{Observation Space.}~~
The observation space is composed as follows:
\begin{equation}
    s_t = \left( l_t, \bar{d}_t, \text{cos}(\varphi^y_t), \text{sin}(\varphi^y_t)\right),
\end{equation}
where $l_t$ is the water level at time $t$, $\bar{d}_t$ is the moving average of past water demands, and $\varphi^y_t \in [0, 2\pi]$ represents the angular position of the current time over the entire year, given by $\varphi^y = \frac{2\pi \tau_y}{T_y}$, where $\tau_y \in [0, T_y]$ is the current time in seconds and $T_y$ is the total number of seconds in a year.

\textbf{Action Space.}~~
The action is a continuous variable $a_t \in \mathbb{R}^+$, representing the amount of water to release per unit of time.

\textbf{Reward Function.}~~
The reward at time $t$ is $r_t = \left[ r_{d}(a_t) + r_{\text{of}}(a_t) + r_{\text{st}}(a_t) \right] + \lambda_1 r_{\text{clip}}(a_t) + \lambda_2 r_{w}(a_t)$, where $r_{d}(a_t)$, $r_{\text{of}}(a_t)$ and $r_{\text{st}}(a_t)$ are the quantities in Equation~\ref{eq:dam_objective}, while $r_{\text{clip}}(a_t)$ and $r_{w}(a_t)$ are two terms designed to discourage actions beyond the physical constraints of the environment and to discourage water releases that are higher than the daily demand, respectively. The two positive hyperparameters $\lambda_1$ and $\lambda_2$ regulate the importance of these two additional penalty terms. The presence of multiple contrastive components enables the development of MORL paradigms. 

\begin{figure}[t]
\centering
\begin{minipage}{.49\textwidth}
    \subfloat[Mean cumulative rewards.]{%
  \includegraphics[width=\linewidth]{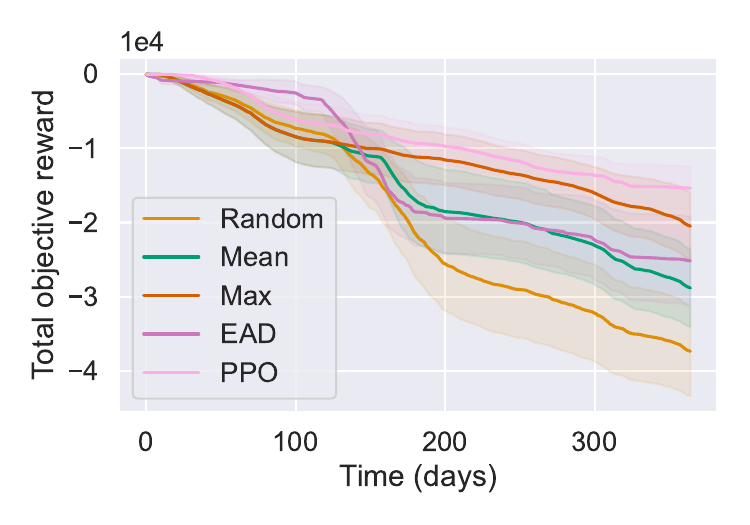} \label{fig:dam_avg_reward}
}
\end{minipage}
\hfill
\begin{minipage}{.49\textwidth}
   \subfloat[Boxplot of returns.]{%
   \includegraphics[width=\linewidth]{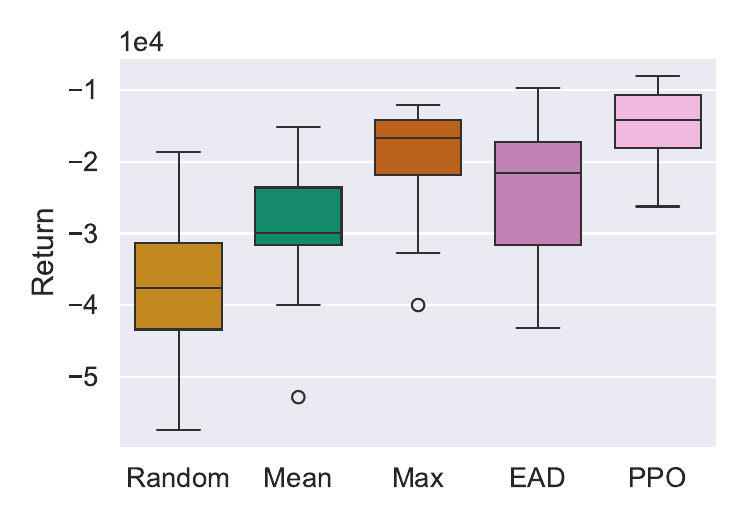} \label{fig:dam_box}
} 
\end{minipage}
\caption{Test performances with confidence intervals on \texttt{DamEnv}. Thirteen different episodes have been considered with a time horizon of one year.}
\end{figure}

\textbf{Benchmarking.}~~
We employed an off-the-shelf implementation of the Proximal-Policy Optimization (PPO)~\cite{schulman2017ppo} algorithm as a benchmark state-of-the-art RL approach for the \texttt{DamEnv} task.
%, as implemented in the SKRL library~\cite{serrano2023skrl}. 
We evaluated the trained agent against four rule-based baselines: the \textit{Random} policy, which selects actions uniformly at random; the \textit{Mean} policy, which selects the mean value of the action space; the \textit{Max} policy, which selects the maximum value of the action space; and the \textit{EAD} policy,  which sets actions based on an exponential moving average of previous demands. The experiments conducted on $13$ test episodes highlight the capability of the PPO agent to perform better than rule-based strategies. In particular, we can observe a better daily control of the PPO agent throughout one year, as shown in Figure~\ref{fig:dam_avg_reward}, and a larger average return with small variability, as highlighted in Figure~\ref{fig:dam_box}. Detailed results show that PPO avoids dam overflows much more effectively than the baselines, as detailed in the Appendix.
%\ref{sec:experiment_details:damenv}.

% IN ALTERNATIVA CON SUBFLOAT

%\begin{figure}[ht]
%\centering
%\begin{minipage}{.49\textwidth}
%  \centering
%  \subfloat[]
%  {\includegraphics[width=\linewidth]{contents/imgs/dam_avg_reward.pdf}
%  %\caption{Mean cumulative rewards.}
%  \label{fig:dam_avg_reward}}\qquad
%\end{minipage}%
%\begin{minipage}{.49\textwidth}
%  \centering
%    \subfloat[]{\includegraphics[width=\linewidth]{contents/imgs/dam_boxplot.pdf} 
%    %\caption{boxplots of the returns.}
%    \label{fig:dam_boxplot}}\qquad
%\end{minipage}
%\caption{Test performances with confidence intervals. Thirteen different episodes have been considered, with a time horizon of one year.}
%\end{figure}

% Boxplots

%% file: contents/envs/elevator.tex
\subsection{\texttt{ElevatorEnv}}
\texttt{ElevatorEnv} is a simplified adaptation of the well-known elevator scheduling problem introduced by~\citet{crites1995elevator}. Similarly to a subsequent work~\cite{yuan2008elevator}, we design a discrete environment that simulates \emph{peak-down traffic}, typical of scenarios such as office buildings at the end of a workday.
In this environment, a single elevator serves a multi-floor building with $F$ floors and is tasked with transporting employees to the ground floor ($f=0$). The episode unfolds over $T$ discrete time steps. At each floor $f \in \{1, \dots, F\}$, new passengers arrive according to a \emph{Poisson process} with rate $\lambda_f$.
%i.e., the number of arrivals per timestep follows is:
%\begin{equation}
%\Pr(N_f = k) = \frac{\lambda_f^k \, e^{-\lambda_f} }{k!}, \quad k = 0, 1, 2, \dots
%\end{equation}

Arriving passengers join a queue on their respective floor, provided the queue length is below a predefined threshold $W_{f,\text{max}}$. Otherwise, they opt to take the stairs.
The \emph{goal} of the elevator controller is to minimize the cumulative \emph{waiting time} of all transported passengers throughout the episode. This can be formalized as minimizing the cost:
\begin{equation}
    \min \sum_{t=1}^{T} \Big(\sum_{f=1}^{F} w_{f,t}+c_t\Big),
\end{equation}
where $w_{f,t}$ denotes the total waiting time of individuals at floor $f$ at time $t$. This setting defines a challenging \emph{load management problem}, involving a trade-off between serving higher floors with longer queues and minimizing elevator travel time. Furthermore, the discrete and restrained formulation of \texttt{ElevatorEnv} facilitates the development of provably efficient RL methods, without losing the connection with the underlying real-world task.

\textbf{Observation Space.}~~
The observation space is structured as follows:
\begin{equation}
    s_t = (h_t, c_t, \mathbf{w}_t, \mathbf{k}_t) \label{eq:elevator-objective},
\end{equation}
where $h_t \in \{0, \dots, H\}$ denotes the vertical position of the elevator within the building at time $t$, being $H$ the maximum reachable height, $c_t \in \{0, \dots, C_{\text{max}}\}$ indicates the current load of the elevator, in number of passengers, up to the maximum capacity $C_{\text{max}}$, 
%$v_t \in \{-v, 0, v\}$ represents the direction and speed of the elevator at time $t$, where $v$ is the unit movement speed (positive for upward, negative for downward, and zero for idle),
and $\mathbf{w}_t \in \mathbb{N}^{F}$ and $\mathbf{k}_t \in \mathbb{N}^{F}$ represent the actual number of people waiting in the queue and the new arrivals at each floor. 

%(excluding the ground one) at time $t$.

\textbf{Action Space.}~~
The action space is defined by the discrete action variable $a_t \in \{\text{\textit{u}}, \text{\textit{d}}, \text{\textit{o}}\}$ which indicates whether the elevator has to move upwards (\textit{u}), move downwards (\textit{d}), or stay stationary and open (\textit{o}) the doors. Actions are mutually exclusive and applied at each time step $t$.

\textbf{Reward Function.}~~
The instantaneous reward is $r_t=-(\sum_f w_{f,t} + c_t) + \mathbbm{1}_{\{c_t = 0\}}\beta \,c_{t-1}$, i.e., at each step $t$ we penalize the presence of individuals, either waiting in queues ($w_{f,t}$) or inside the elevator ($c_t$), as in Equation~\eqref{eq:elevator-objective}. In addition, we grant a positive reward when passengers are successfully delivered to the ground floor, i.e., when the elevator becomes empty. The positive hyperparameter $\beta > 0$ controls the reward magnitude for offloading $c_{t-1}$ passengers.

\begin{figure}[t]
\centering
\begin{minipage}{.49\textwidth}
  \subfloat[Mean cumulative rewards.] 
  {\includegraphics[width=\linewidth]{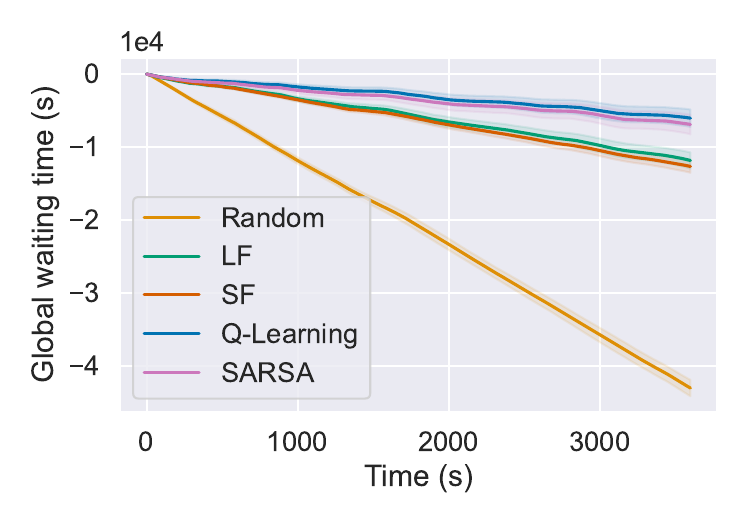} \label{fig:elevator-avg_reward}} 
\end{minipage}%
\hfill
\begin{minipage}{.49\textwidth}
    \subfloat[Boxplot of returns.]{\includegraphics[width=\linewidth]{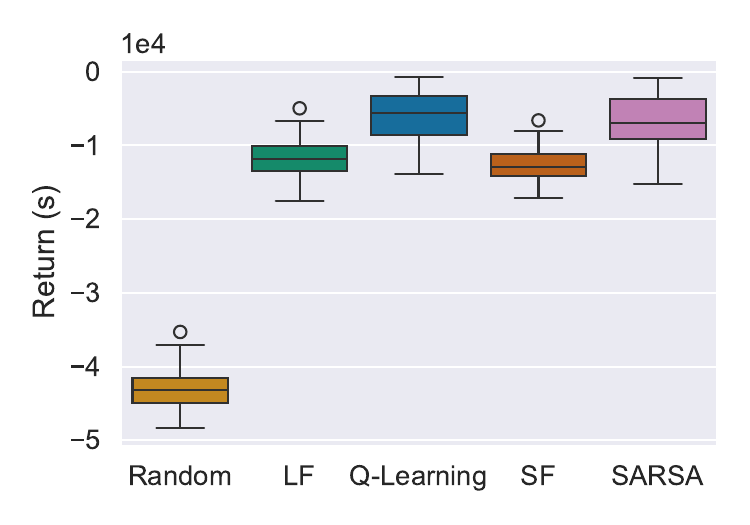} \label{fig:elevator-boxplot}}\qquad
\end{minipage}
\caption{Performance of baselines in terms of mean cumulative reward (a) and average return (b) on \texttt{ElevatorEnv}. Results collected over $30$ different episodes.}
\end{figure}

\textbf{Benchmarking.}~~
For the \texttt{ElevatorEnv} task, we adopt two well-known tabular RL algorithms: Q-Learning~\cite{watkins1992q} and SARSA~\cite{Sutton1998}. Such methods are evaluated against different rule-based strategies, i.e., the \emph{Random} policy, and the \textit{Longest-First} (LF) and the \textit{Shortest-First} (SF) policies, which prioritize the floor with a higher or lower number of waiting people, respectively. As shown in Figure~\ref{fig:elevator-avg_reward}, both RL algorithms consistently outperform the other rule-based solutions, considerably reducing the global waiting time. In particular, as reported in Figure~\ref{fig:elevator-boxplot}, Q-Learning shows higher performance than SARSA, which, due to its inherent nature, tends to play more conservative actions.

%% file: contents/envs/microgrid.tex
\subsection{\texttt{MicrogridEnv}}
\texttt{MicrogridEnv} simulates the operation of a microgrid within the context of electrical power systems. Microgrids are decentralized components of the main power grid that can function either in synchronization or in islanded mode. 
%They typically consist of multiple energy sources (both renewable and non-renewable), battery energy storage systems (BESSs), and various energy-consuming entities. 
In this scenario, the control point is placed on the battery component, which must find the best strategy to manage the accumulated energy over time optimally. Formally, the controller wants to maximize its total profit over a time horizon of $T$. 
%following a specific policy, $\pi = (a_0, \ldots, a_{T})$. 
Hence, the objective is:
%to find the optimal policy $\pi^*$ so that:
\begin{equation}
    \max \sum_{t=1}^T \ [r_{\text{trad}}(a_t) + r_{\text{deg}}(a_t)] \label{eq:mg_objective},
\end{equation}
where $r_{\text{trad}}(a_t) \in \mathbb{R}$ is the reward/cost gained from the exchanges of energy with the market, and $r_{\text{deg}}(a_t) < 0$ is the cost due to battery degradation.
The benchmark leverages real-world datasets, as detailed in the Appendix, and the battery behavior is modeled using a digital twin of a BESS~\cite{salaorni2023ernesto}. Each episode is formulated as an infinite-horizon problem and terminates either when the dataset is exhausted or the battery reaches its end-of-life condition. 
%The datasets have an hourly resolution, and control actions are applied at the same temporal granularity.
Moreover, the presence of energy market trends allows the usage of \texttt{MicrogridEnv} for frequency adaptation analysis.

\textbf{Observation Space.}~~
The observation space comprises variables regarding the internal state of the system and uncontrollable signals received from the environment. Formally:
\begin{equation}
    s_t = \big( \sigma_t, K_t, \widehat{P}_{D,t}, \widehat{P}_{G,t}, p_t^{\text{buy}}, p^{\text{sell}}_t, \text{cos}(\varphi^d_t), \text{sin}(\varphi^d_t), \text{cos}(\varphi^y_t), \text{sin}(\varphi^y_t) \big),
\end{equation}
where $\sigma_t$ is the storage state of charge, $K_t$ is the battery temperature, $\widehat{P}_{D,t}$ is the estimate of energy demand $P_{D,t}$, $\widehat{P}_{G,t}$ is the estimate of energy generation $P_{G,t}$, $p^{\text{buy}}_t$ and $p^{\text{sell}}_t$ are the buying and selling energy market prices, respectively, $\varphi^d_t \in [0, 2\pi] $ is the angular position of the clock in a day, and $\varphi^y_t \in [0, 2\pi]$ is the angular position of the time over the entire year.

\textbf{Action Space.}~~
The action space is determined by the continuous action variable $a_t \in [0, 1]$, representing the proportion of energy to \textit{dispatch} (\textit{take}) to (from) the BESS. The action operates with the net power computed as $P_{N,t} = P_{G,t} - P_{D,t}$. If $P_{N,t} > 0$, it regulates the proportion of energy used to charge the battery or sold to the main grid. Conversely, if $P_{N,t} < 0$, the action balances the proportion of energy taken from the energy storage or bought from the market.

\textbf{Reward Function.}~~
The instantaneous reward is 
$
r_t = [r_{\text{trad}}(a_t) + r_{\text{deg}}(a_t)] + \lambda r_{\text{clip}}(a_t) \label{eq:mg_reward}
$,
where $r_{\text{clip}}(a_t)$ is a penalty that discourages actions that do not respect physical constraints, weighted by the hyperparameter $\lambda$.
The first two elements, instead, are the same components of the objective function in Equation~\eqref{eq:mg_objective}, whose contrastive optimization enables multi-objective RL approaches.

\begin{figure}[t]
\centering
\begin{minipage}{.49\textwidth}
  \centering
  \subfloat[Mean cumulative rewards.] 
  {\includegraphics[width=\linewidth]{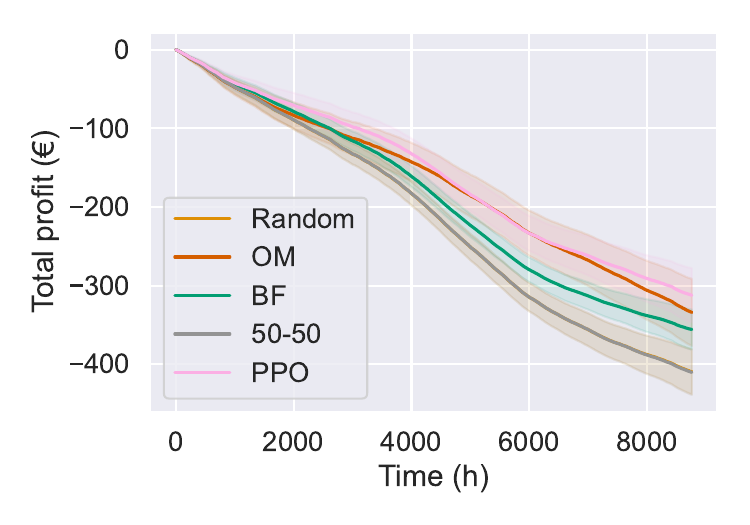} \label{fig:mg-avg_reward}} 
\end{minipage}%
\hfill
\begin{minipage}{.49\textwidth}
  \centering
    \subfloat[Boxplot of returns.]{\includegraphics[width=\linewidth]{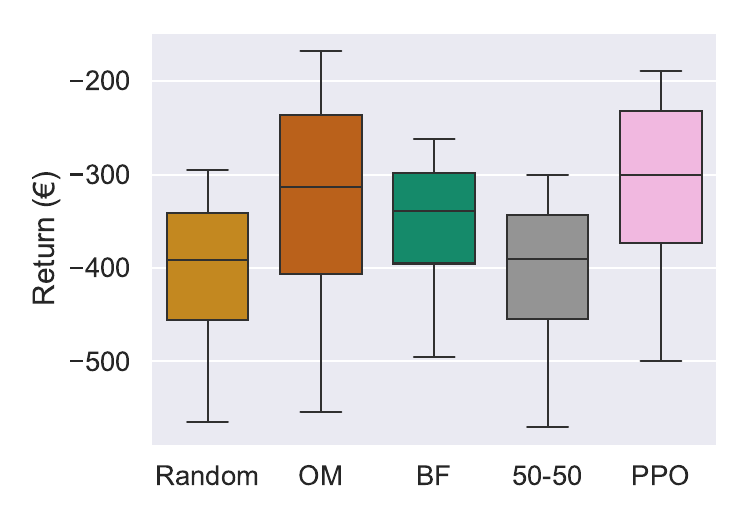} \label{fig:mg-boxplot}}\qquad
\end{minipage}
\caption{Performance of baselines in terms of mean cumulative reward (a) and average return (b) on \texttt{MicrogridEnv}. Results have been collected over $28$ different episodes.}
\end{figure}

\textbf{Benchmarking.}~~
For the \texttt{MicrogridEnv}, we compare an RL agent trained with PPO~\cite{schulman2017ppo} against several rule-based policies: %Among these we have: 
the \textit{Random} policy; the \textit{Only-market} (OM) policy, which forces the interaction with the grid without using the battery; the \textit{Battery-first} (BF) policy, which fosters the battery usage; and the \textit{50-50} policy, which adopts a behavior in the middle between OM and BF. Figure~\ref{fig:mg-avg_reward} shows that, during testing, PPO achieves higher profit than rule-based strategies. However, as reported in Figure~\ref{fig:mg-boxplot}, PPO has a large variance, suggesting the need for novel RL algorithms to achieve more consistent behavior.

%% file: contents/envs/robofeeder.tex
\subsection{\texttt{RoboFeederEnv}} \label{sec:robofeeder}
\texttt{RoboFeederEnv} is a collection of environments designed to pick small objects from a workspace area with a 6-degree-of-freedom (6-DOF) robotic arm. This task involves two primary challenges: determining the \textit{picking order} of the objects and identifying the precise \textit{grasping point} on each object for successful pickup and placement. To closely mimic the behavior of the commercial robotic system, a simulation emphasizing contact interactions is conducted using MuJoCo~\cite{todorov2012mujoco}. 
%The simulator models both the robot’s physical dynamics and its interactions with objects via the end-effector. 
This environment supports goal-oriented training, enabling the robot to learn how to identify the appropriate grasping points and, more broadly, to determine the most efficient order of picking.
%in which to pick up objects based on their assigned priorities. 
Unlike most robotic simulators, \texttt{RoboFeederEnv} is uniquely tailored 
%for training algorithms that 
to operate at the trajectory planning level rather than through low-level joint control, which is more realistic in industrial applications, given the impossibility of accessing and modifying proprietary kinematic controllers.

%This distinction reflects a key real-world constraint: commercial 6-DOF robotic arms typically rely on proprietary kinematic controllers for joint-level operations, which are not directly accessible or modifiable. By focusing on high-level planning, \texttt{RoboFeederEnv} more accurately mirrors the real-world development pipeline, where robot programmers interact primarily through trajectory commands rather than raw joint control.

Due to the hierarchical nature of the problem, we split the setting into two underlying environments: \texttt{RoboFeeder-picking} and \texttt{RoboFeeder-planning}.

\subsubsection{\texttt{RoboFeeder-picking}}
\texttt{Gym4ReaL} includes two types of picking environments of increasing difficulty:
\begin{itemize}[leftmargin=*, noitemsep, topsep=-1pt]
    \item \texttt{picking-v0}: a simpler environment where the top-down image is pre-processed by cropping around detected objects, %, based on the output of a prior object detection module. 
     reducing the complexity of the visual input, thus of the observation space;
    \item \texttt{picking-v1}: a more challenging environment where the observation is the full camera image. %, without any object detection pre-processing.
    % in which the agent must learn directly from 
\end{itemize}

\textbf{Observation Space.}~~ 
%Since the environment employs visual inputs, t  space
The observation is defined by the visual input $s_t = \mathbf{X}_t\in \mathbb{R}^{H \times W \times C}$, 
%$\mathcal{S} = \mathbb{R}^{1 \times H \times W \times C}$, 
where each image $\mathbf{X}_t$ is represented by a tensor of height $H$, width $W$, and channel $C$, and is captured by a camera positioned on top of the working area. 
%The camera capturing the image is positioned on top of the working area, with a top-down perspective. 
%Indeed, as in real-world systems, the robot is designed to be driven by a vision system. 
Within the \texttt{picking-v0} environment,  the image tensor is restricted to $\widehat{\mathbf{X}}_t \in \mathbb{R}^{\widehat{H} \times \widehat{W} \times C}$, with $\widehat{H}$ and $\widehat{W}$ cropped image dimensions. 

\textbf{Action Space.}~~ 
The action space is determined by the continuous action $a_t = (x_t,y_t)$, 
%as
%$
%\mathcal{A} = \{(x, y) \mid x, y \in \mathbb{R}\}
%$
where $(x_t, y_t)$ are relative coordinates within the segmented image, corresponding to the target grasping point.

\textbf{Reward Function.}~~
The reward function is designed to foster successful object picking while penalizing unfeasible or suboptimal actions. Formally, the instantaneous reward is $r_t = 1$ if the object is correctly picked up, $r_t = -1$ if the action is unfeasible, or $r_t = -1 + r_{d,t} +r_{\theta, t}$ otherwise, where $r_{d,t}$ is a distance-based shaping term that rewards proximity of the end-effector to the object, and $r_{\theta, t}$ is a rotation-based shaping term that incentivizes alignment with the desired grasping orientation.

\textbf{Benchmarking.}
We evaluate the performance of a trained PPO~\cite{schulman2017ppo} agent against a fixed action rule-based strategy on the \texttt{picking-v0} environment. The task involves objects uniformly distributed within the workspace, requiring non-trivial generalization capabilities. Figures~\ref{fig:picking-baseline} and~\ref{fig:picking-ppo} report how the baseline exhibits consistently poor performance, while the PPO agent achieves higher and more evenly distributed success rates, highlighting its capability to learn an effective picking strategy.

\iffalse %COMMENTED
\begin{figure}
\centering
\begin{minipage}[H]{.45\textwidth}
  \subfloat[Baseline.] 
  {\includegraphics[width=\linewidth]{contents/imgs/heatmap_overlay_baseline.png} \label{fig:picking-baseline}}\qquad 
\end{minipage}%
\begin{minipage}[H]{.45\textwidth}
    \subfloat[PPO.]{\includegraphics[width=\linewidth]{contents/imgs/heatmap_overlay_ppo.png} \label{fig:picking-ppo}}\qquad
\end{minipage}
\caption{Picking success rate across the entire workspace.}
\end{figure}
\fi

\subsubsection{\texttt{RoboFeeder-planning}}
The \texttt{RoboFeeder-planning} is an environment aiming to decide the order to follow for picking the objects in the work area. It is a high-level task w.r.t. \texttt{RoboFeeder-picking}, not involving the direct control of the robot, but only concerning the optimal picking schedule. 

\textbf{Observation Space.}~~ 
The observation space is defined by the vector of visual input $s_t = [\mathbf{X}_{1,t}, \dots, \mathbf{X}_{N,t}]$, with $\mathbf{X}_{i,t} \in \mathbb{R}^{H \times W \times C}$, where $N$ is the maximum number of images that can be processed and $\mathbf{X}_{i,t}$ is an image defined as in the \texttt{picking-v0} task.
%and $H$, $W$, and $C$ are the height, width, and number of channels of the normalized image patches, respectively. 
Each of the $N$ image patches corresponds to a cropped and scaled region of a detected object.

\textbf{Action Space.}~~ 
The action space is determined by the discrete action $a_t \in \{0, 1, \dots, N\}$, selecting the image from $1$ to $N$ containing the object to pick. 
%$1$ to $N$ corresponding to selecting one of the $N$ candidate objects for grasping. 
Action $0$, instead, is a special \emph{idle} action that can be chosen when no graspable objects are available. This formulation enables continuous deployment since the robot can remain idle while waiting for the arrival of new objects.

\textbf{Reward Function.}~~ 
The immediate reward is $r_t=1$, if the selected object is correctly picked, $r_t=-1$ if it is not picked, and $r_t = - \sum_{i=1}^{M} \mathbbm{1}_{\{\text{obj}_i \text{ not picked but graspable}\}}$ if the agent plays the \textit{idle} action $a_t = 0$ while graspable objects are present, with $M$ being the currently available objects.

\begin{figure}[t]
\centering
\begin{minipage}{.325\textwidth}
  \subfloat[\texttt{picking-v0}: baseline.] 
  {\includegraphics[width=\linewidth]{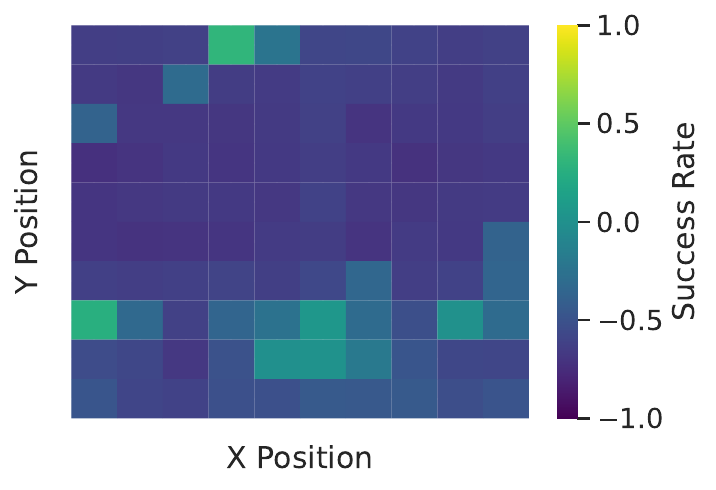} \label{fig:picking-baseline}} 
\end{minipage}
\hfill
\begin{minipage}{.325\textwidth}
    \subfloat[\texttt{picking-v0}: PPO.]{\includegraphics[width=\linewidth]{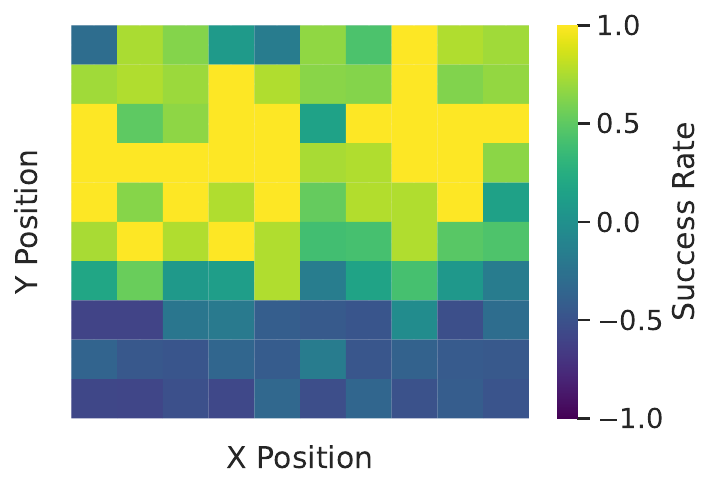} \label{fig:picking-ppo}}
\end{minipage}
\hfill
\begin{minipage}{0.32\textwidth}
        \subfloat[\texttt{RoboFeeder-planner}.]{
        \includegraphics[width=\linewidth]{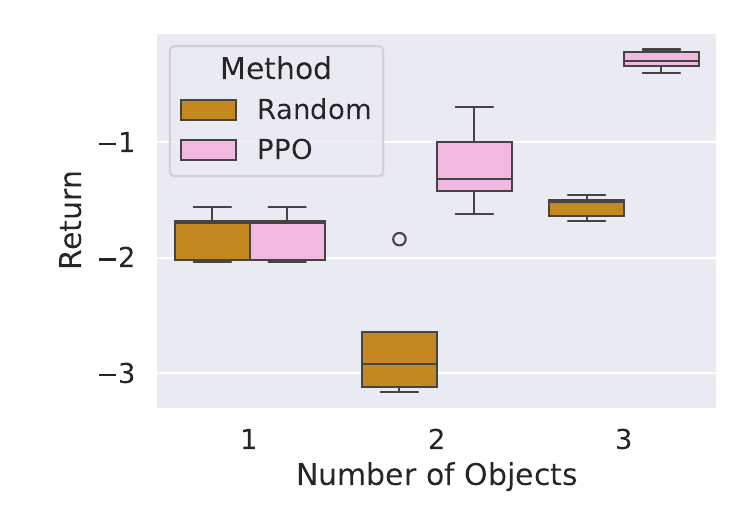}
        \label{fig:planning_baseline}} 
    \end{minipage}
\caption{Heatmap of the success rate of picking tasks across the entire workspace with baseline (a) and PPO (b) (the higher, the better). Comparison between \textit{Random} policy and PPO within the planning problem (c) (average return over $50$ episodes and $5$ different random seeds).}
\end{figure}

\textbf{Benchmarking.}~~
In Figure~\ref{fig:planning_baseline}, we compare the efficiency of a trained PPO~\cite{schulman2017ppo} agent against a \textit{Random} strategy. Results highlight the agent's capability to determine an optimal picking schedule by distinguishing objects placed in a favorable position to be picked up. Moreover, as the number of objects increases, the gap between the average return of PPO and the baseline increases too.

%% file: contents/envs/trading.tex
\subsection{\texttt{TradingEnv}}
\label{sec:tradingenv}
\texttt{TradingEnv} provides a simulated market environment, trained with historical foreign exchange (Forex) data relative to the EUR/USD currency pair, where the objective is to learn a profitable intraday strategy.
%, obtained from \url{https://www.histdata.com}), 
The problem is framed as episodic: each episode starts at 8:00 EST and ends at 18:00 EST when the position must be closed.
At each step, based on its expectations, the agent can open a \textit{long} position (i.e., buy a fixed amount of the asset), remain \textit{flat} (i.e., take no action), or open a \textit{short} position (i.e., short sell a fixed amount of the asset).
%a fixed amount of the asset it does not own, intending to repurchase it later at a lower price to generate profit).
%in these settings
Typical baselines include passive strategies, such as \textit{Buy\&Hold} (B\&H) and \textit{Sell\&Hold} (S\&H), which consist of maintaining fixed positions.
%In Table~\ref{tab:benchmark_features}, this environment includes different characteristics that make it difficult. 

Trading tasks are typically subjected to several challenges.
For example, the state has to be carefully designed to deal with the low signal-to-noise ratio, and it is typically large-dimensional, including past prices and temporal information. 
%Since all the information included in the state to represent the market state is obtained from historical data, 
Moreover, the environment is partially observable, and financial markets are non-stationary.
%Financial markets are well-known to be non-stationary and typically show different market regimes.
%Related to the RL paradigms of the Table~\ref{tab:benchmark_features}, 
%A promising approach is the use of frequency adaptation methods, employed to calibrate 
Another relevant aspect is the calibration of the trading frequency, considering the amount of noise and transaction costs.
%is associated with the trading frequency that is typically calibrated considering the amount of noise and transaction costs. 
In addition, risk-aversion approaches can be of interest, considering not only the profit-and-loss (P\&L) but also the variance among episodes.

\textbf{Observation Space.}~~
The observation space is composed of two components: \textit{market state} and \textit{agent state}. 
The \textit{market state} 
%includes all the features that are useful to represent the market. In the basic version, it 
% during the last hour
includes calendar features and recent price variations, namely the last $60$ delta mid-prices, where a delta mid-price is defined as $d_{k,t} = \frac{p_{t-k} - p_{t-k-1}}{p_{t-k-1}}$, with $k \in \{0,\dots,59\}$. 
%To the \textit{market state} belongs also the minute of the day.
The \textit{agent state} component, on the contrary, includes the current position $z_t$, that is, the action that was previously played. 
%and it is affected by previous actions (e.g., at 8:00 EST, the agent starts flat until it decides to open a long or short position. Then, the position is kept until it decides to change the position to either flat or short.). 
Formally, the state in this setting is:
\begin{equation}
    s_t = \big( \mathbf{d}_t, \text{cos}(\varphi^{day}_t), \text{sin}(\varphi^{day}_t), z_t \big),
\end{equation}
where $\mathbf{d}_t = [d_{0,t}, \dots, d_{59,t}]$ is the vector of the last $60$ delta mid prices at time $t$, $\varphi^{day}_t \in [0, 2\pi]$ is the angular position of the current time over the trading period, and $z_t=a_{t-1}$ is the agent position. 
%with $K=59$ 

\textbf{Action Space.}~~
The action space is determined by a discrete variable $a_t \in \{\textit{s}, \textit{f}, \textit{l}\}$, where \emph{s} (\textit{short}) indicates that the agent is betting against EUR, supposing a decline in the value relative to USD; \emph{f} (\textit{flat}) indicates no market exposition; and \emph{l} (\textit{long}) means that the agent expects that the relative EUR value will increase. Each action refers to a fixed amount of capital $C$ to trade.
%and corresponds to a fixed amount of capital C (in the experiments, we set C = €100k).

\textbf{Reward Function.}~~
The immediate reward at time $t$ is the signal $r_{t} = a_{t-1}(p_{t} - p_{t-1}) - \lambda|a_t - z_t|$, where 
%\begin{equation}
%    R_{t+1} = a_t(price_{t+1} - price_t) - f_t|a_t - a{t-1}|
%\end{equation}
the first term is related to the P\&L obtained from a price change, and the second component regards the commissions paid when the agent changes its position, being $\lambda$, a constant transaction fee.

\begin{figure}[t]
    \begin{minipage}[t]
    {0.45\linewidth}
        \subfloat[P\&L curve on test set.]{
         \centering
        \includegraphics[width=\linewidth]{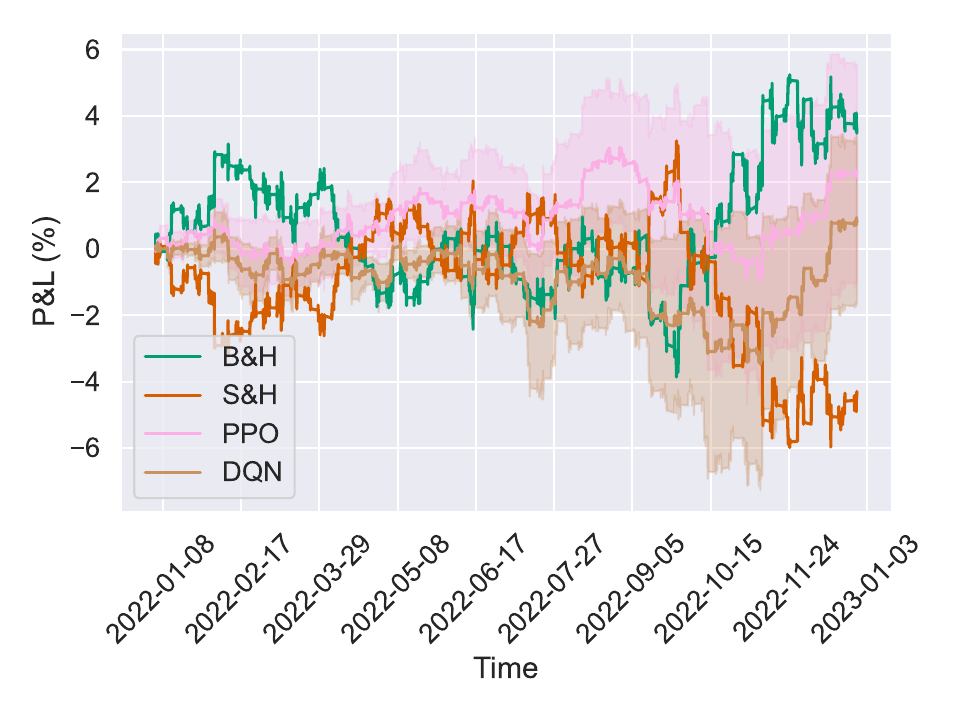} 
    \label{fig:tradingenv_PPODQNvsBaselines:test}
        }
    \end{minipage}
    \hfill
     \begin{minipage}[t]
     {0.49\linewidth}
     \vspace*{-0.5cm}
     \subfloat[Boxplot of P\&L on test set.]{
        \includegraphics[width=\linewidth]{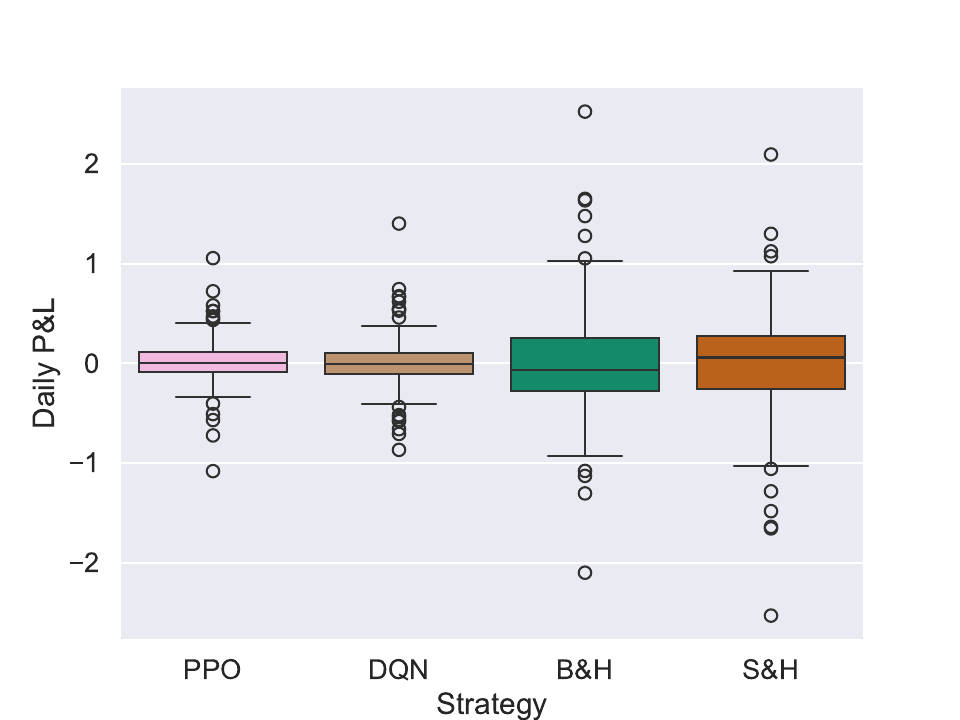}
    \label{fig:tradingenv_PPODQNvsBaselines:boxplot}
        }
    
    \end{minipage}
     
    \caption{Performances of PPO and DQN against baselines B\&H and S\&H on Test (a) Daily Performance on Test (b) on \texttt{TradingEnv}. Mean and standard deviation are computed over 6 seeds.}
\end{figure}

\textbf{Benchmarking.}~~
We trained agents using off-the-shelf implementations of PPO~\cite{schulman2017ppo} and DQN~\citep{Mnih2015Atari} on \texttt{TradingEnv}. Their performance against common passive baselines, B\&H and S\&H, are evaluated on a test year (Figure~\ref{fig:tradingenv_PPODQNvsBaselines:test}). As expected, neither PPO nor DQN is able to consistently outperform the baselines, due to the complexity of the problem. However, RL remains a valid candidate to tackle trading tasks, as it significantly reduces the daily variability of the P\&L (Figure~\ref{fig:tradingenv_PPODQNvsBaselines:boxplot}).

%\paragraph{Limitations.}
%The main limitations of this environment are related to the quality of market data, that typically are very expensive and it is difficult to find for free. The Forex data used in this environment contains temporal gaps which are addressed by both removing days with large number of gaps and by applying carefully forward-filling strategies, minimizing the risk of data leakages.
%Moreover, in this environment we are not taking into account the market impact of the action. Typically, in high liquid markets as the EUR/USD currency pair and with reasonable order size this can be neglected, but must be taken into consideration in general.

%% file: contents/envs/wds.tex
\subsection{\texttt{WaterDistributionSystemEnv}}
\texttt{WaterDistributionSystemEnv} simulates the evolution of a hydraulic network in charge of dispatching water across a residential town. A network is composed of different entities, such as storage tanks, pumps, pipes, junctions, and reservoirs, and the main objective of the system is the safety of the network. To achieve such a goal, we have to ensure optimal management of hydraulic pumps, which are in charge of deciding how much water should be collected from reservoirs and dispatched to the network. The pumps' controller must guarantee network resilience by maximizing the demand satisfaction ratio (DSR) while minimizing the risk of overflow. Formally, the objective is
% learn a strategy $\pi^*$ to 
% \pi^* = \argmax_\pi 
\begin{equation}
    \max \sum_{t=1}^T \ [r_{\text{DSR}}(a_t) + r_{\text{of}}(a_t)] \label{eq:wds_objective},
\end{equation}
where $r_{\text{DSR}}(a_t) \in [0,1]$ is the ratio between the supplied demand on the expected demand at time $t$, and $r_{\text{of}}(a_t) \in [0,1]$ is a normalized penalty associated with the tanks' overflow risk. 

The environment leverages the hydraulic analysis framework Epanet~\cite{epanet2000}, which provides the mathematical solver for water network evolution, and realistic datasets of demand profiles. Therefore, \texttt{WDSEnv} may also be suitable to test imitation learning methods, having at disposal an expert policy from the \textit{.inp} configuration file of networks read by Epanet.

\textbf{Observation Space.}~~
The observation space includes the internal state of the network and an estimation of the global demand profile that the system is asked to deal with. Formally:
\begin{equation}
    s_t = \left(\mathbf{h}_t, \mathbf{p}_t, \widehat{d}_t,  \text{cos}(\varphi^d_t), \text{sin}(\varphi^d_t) \right),
\end{equation}
where $\mathbf{h}_t \in \mathbb{R}^{L}$ is the vector of $L$ tank levels at time $t$, $\mathbf{p}_t \in \mathbb{R}^{J}$ is the vector of $J$ junction pressures at time $t$, $\widehat{d}_t$ is the estimated total demand at time $t$, and $\varphi^d_t \in [0, 2\pi] $ is the angular position of the clock in a day. Finally, although all tanks must be monitored, we can reduce the dimensionality of the observation space by considering only junctions placed in strategic positions.

\textbf{Action Space.}~~ 
The discrete action variable $a_t \in \mathbb{N}$ can assume values in $\{0, \dots, 2^{P}-1\}$, with $P$ number of pumps within the system. The action determines the combination of open/closed pumps.  

\textbf{Reward Function.}~~
The instantaneous reward given by the environment is 
$
r_t = r_{\text{DSR},t}(a_t) + r_{\text{of},t}(a_t)
$,
where the terms are those described in the objective function in Equation~\eqref{eq:wds_objective}.

%\textbf{Alternative version}~~
%An alternative environment built upon \texttt{WDSEnv} and conceived from the high-fidelity simulator DHALSIM~\cite{murillo2023high-fidelityI, murillo2023high-fidelityII}, treats the hydraulic network as a cyber-physical system (CPS). Indeed, such an environment simulates PLC sensors and actuators used to read measurements of the network and apply modifications to the pumps' status. The introduction of this communication layer allows the simulation of cyberattacks, such as man-in-the-middle (MITM) or denial of service (DOS), hindering the information exchange. In particular, these attacks can obstruct the operation and learning of the agent, which could receive altered signals or perform actions that will not be applied. In this task, the observation space includes a boolean variable indicating the presence of ongoing attacks, simulating the presence of an omniscient attack detection system.

\begin{figure}[t]
\centering
\begin{minipage}{.49\textwidth}
  \centering
  \subfloat[Mean cumulative reward.] 
  {\includegraphics[width=\linewidth]{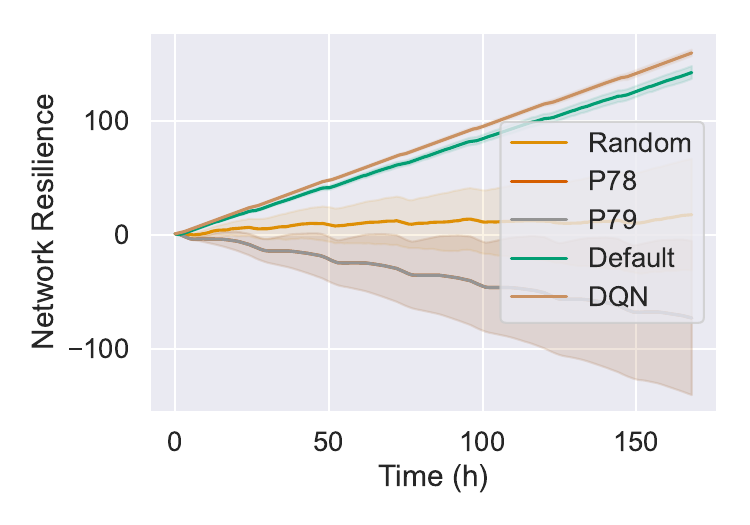} \label{fig:wds-avg_reward}} 
\end{minipage}%
\hfill
\begin{minipage}{.49\textwidth}
  \centering
    \subfloat[Boxplot of returns.]{\includegraphics[width=\linewidth]{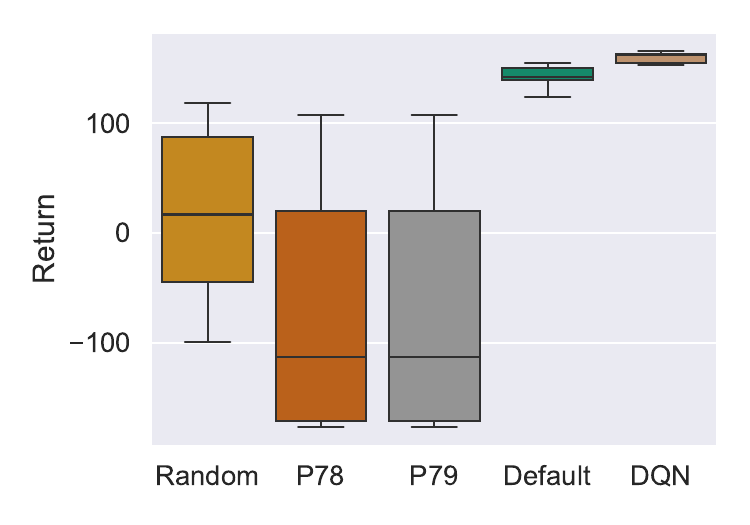} \label{fig:wds-boxplot}}\qquad
\end{minipage}
\caption{Performance of baselines in terms of mean cumulative resilience (a) and average return (b) on \texttt{WDSEnv}. Results have been collected over $20$ different episodes.}
\end{figure}

\textbf{Benchmarking.}~~
The \texttt{WDSEnv} is benchmarked adopting DQN~\cite{Mnih2015Atari}, which is compared with different rule-based baselines: the \textit{Random} policy, \textit{P78} and \textit{P79} policies, which act by keeping active only the relative pump (namely P78 or P79, respectively), and the \textit{Default} policy, which executes the default control rules contained within the \textit{.inp} configuration file of the network, changing the control action depending on the current tank level. As depicted in Figure~\ref{fig:wds-avg_reward}, DQN achieves a higher level of resilience with respect to other baselines. Moreover, Figure~\ref{fig:wds-boxplot} shows that it has a more consistent behavior and low variance, a crucial characteristic for the resilience and safety of the water network.

%% file: contents/conclusions.tex
\section{Discussion and Conclusion} \label{sec:conclusions}
In this work, we presented \texttt{Gym4ReaL}, a benchmarking suite designed to model several real-world environments specifically tailored for RL algorithms. 
Unlike standard benchmarking suites, which often rely on idealized tasks, \texttt{Gym4ReaL} represents a novel library that allows for evaluating new RL methods in realistic playgrounds. Notably, the \texttt{Gym4ReaL} suite includes environments designed to capture common real-world challenges, such as limited data availability, realistic assumptions about physical process dynamics, and constrained exploration, fostering research toward broader adoption of RL methods in practical applications. For this reason, we selected a pool of environments incorporating different \textit{characteristics} and addressing various \textit{RL paradigms}, as shown in Table~\ref{tab:benchmark_features}.

% However, We convey that our work includes some limitations. In particular, we are not addressing multi-agent problems, which are typical of real-world tasks, nor multi-task environments. At the same time, not all the existent \textit{RL paradigms} are easily tailored to our settings, since their application highly depend on the specific task's implementation.

Given the standardized and flexible interface offered by our suite, a broader range of real-world problems and challenges could be easily integrated into our framework. We believe that a collective effort from the RL community can significantly advance the development of realistic, impactful benchmarks. Hence, we encourage researchers and practitioners to explore, contribute to, and adopt \texttt{Gym4ReaL} to evaluate RL algorithms in real-world scenarios.
% warmly
% as a go-to suite for evaluating

%% file: appendix/metadata.tex
\section{Metadata}
\subsection{Hosting and Maintenance}
\texttt{Gym4ReaL} is distributed as a Python package and is publicly available on both GitHub and PyPI. The library follows semantic versioning, with the version associated with this publication labeled as \texttt{v1.0}. A contributor guide is provided in the GitHub repository to support community involvement. The following authors serve as maintainers for the individual environments included in \texttt{Gym4ReaL}:
\begin{itemize}
    \item Davide Salaorni (davide.salaorni@polimi.it): \texttt{ElevatorEnv}, \texttt{MicrogridEnv}, \texttt{WDSEnv}
    \item Vincenzo De Paola (vincenzo.depaola@polimi.it): \texttt{RoboFeederEnv}
    \item Samuele Delpero (samuele.delpero@mail.polimi.it): \texttt{DamEnv}
    \item Giovanni Dispoto (giovanni.dispoto@polimi.it): \texttt{TradingEnv}.
\end{itemize}

\subsection{Licenses and Responsibility}
\texttt{Gym4ReaL} as a whole is released under a Apache-2.0 license~\footnote{\url{https://www.apache.org/licenses/LICENSE-2.0}}.
The authors bear all responsibility in case of violation of rights. \texttt{Gym4ReaL} does not contain any personally identifiable data, nor any offensive content.

\subsection{Reproducibility}
The GitHub repository includes tutorial notebooks demonstrating how to train off-the-shelf RL agents on the various environments, as well as the notebooks used to generate the results we present in this work. Additionally, detailed installation and usage instructions for each notebook are available in the \texttt{Gym4ReaL} wiki: \url{https://daveonwave.github.io/gym4ReaL/}.

\subsection{Intended Usage}
\texttt{Gym4ReaL} is designed as a benchmarking suite for evaluating RL algorithms across a diverse set of real-world-inspired tasks. While substantial effort has been made to ensure that the environments closely reflect the dynamics and constraints of real-world systems, strong performance in \texttt{Gym4ReaL} does not necessarily guarantee equivalent performance in actual real-world deployments.

\subsection{Software and Hardware Requirements}
\texttt{Gym4ReaL} is implemented in Python 3.12. Each environment may require specific dependencies, which are listed in the documentation and provided in the corresponding requirements files.

\texttt{Gym4ReaL} does not require a specific hardware configuration and is portable across all major commercial operating systems. An exception is \texttt{WDSEnv}, which relies on the EPANET simulator since the version that we adopted is not compatible with Apple Silicon processors and requires an Intel-compatible environment. For macOS users, we recommend setting up a dedicated virtual environment to ensure compatibility.

All reported experiments were conducted on an Apple M1 chip (8-core CPU, 8GB of RAM). The per-episode running time for each environment is detailed in Table~\ref{tab:episode_time}. As can be observed, the \texttt{WDSEnv} is computationally expensive, and this is due to the inherent design
of the underlying simulator, which was not originally intended for real-time control or RL applications. We further discuss this in \texttt{WDSEnv} Section~\ref{app:wds-details}.

\renewcommand{\arraystretch}{1.3} % Adjust row spacing
\begin{table}[ht]
    \centering
    \caption{Running time required per episode for the different environments.}
    \vspace{0.2cm}
    \rowcolors{2}{gray!10}{white} % Alternate row colors
    \begin{tabular}{l c}
        \rowcolor{gray!30}
        \textbf{Environment} & \textbf{Time per Episode (s)} \\
        \texttt{DamEnv}       & $0.045$ \\
        \texttt{ElevatorEnv}        & $0.035$ \\
        \texttt{MicroGridEnv}        & $2.781$ \\
        \texttt{RoboFeeder-picking-v0}   &  $0.266$ \\
        \texttt{RoboFeeder-picking-v1}   &  $0.207$ \\
        \texttt{RoboFeeder-planning-v0}   &  $0.076$ \\
        \texttt{TradingEnv}      & $0.007$ \\
        \texttt{WDSEnv}      & $72.932$ \\
    \end{tabular}
    \label{tab:episode_time}
\end{table}

%% file: appendix/deep_dive_table.tex
\section{\textit{Characteristics} and \textit{RL paradigms}}\label{sec:deep_dive_table}

In this section, we define and analyze in detail the environment-specific categories identified in Table~\ref{tab:benchmark_features}.
The table is divided into two main sections:
\begin{enumerate}
    \item \textbf{Characteristics.}~~In the first part of the table, we report the key features of each environment with respect to the standard RL taxonomy. These \emph{Characteristics} are inherent to the design and implementation of the environments provided by \texttt{Gym4ReaL}.
    \item \textbf{RL Paradigms.}~~This second part of the table describes the RL subfields that can be associated with each environment. Specifically, although the benchmarking results provided in the main paper employ standard RL approaches, the environments can be adapted to test novel algorithms related to the respective \emph{RL paradigms} mentioned in the table.
\end{enumerate}

We now elaborate on each of the \emph{Characteristics} and \emph{RL paradigms} reported in the table.

\textbf{Characteristics.}
\begin{itemize}
    \item \textbf{Continuous States.}~~ The environments match this characteristic if the associated state space is continuous. Among the environments provided in Gym4ReaL, 5 out of 6 satisfy this feature, thus reflecting their realism. A notable exception is represented by \texttt{ElevatorEnv}, which allows testing algorithms devised for environments with finite state spaces. 
    \item \textbf{Continuous Actions.}~~This characteristic is matched by the environments when the associated action space is continuous. \texttt{Gym4ReaL} environments cover both the finite and continuous cases.
    \item \textbf{Partially Observable.}~~This characteristic is associated with environments where only partial information is available. Partial information can be related to the difficulty of modeling or accessing relevant state features. This scenario typically happens in trading contexts, such as the one modeled in the proposed \texttt{TradinvEnv}.
    \item \textbf{Partially Controllable.}~~
    %This characteristic reflects scenarios in which, together with the agent's actions, exogenous factors have an impact on the dynamics of the environments. 
    This characteristic refers to environments where part of the state evolves due to external, uncontrollable dynamics. These components, present in the observation space, are independent of the agent’s actions.
    Examples are precipitation in a water control system (\texttt{DamEnv}) or market price variations (\texttt{MicrogridEnv}, \texttt{TradingEnv}).
    \item \textbf{Non-Stationary.}~~This characteristic refers to an (abrupt or gradual) variation of the underlying data distribution, which introduces additional challenges caused by the need for continuous adaptation. This feature is intrinsically related to trading contexts, such as \texttt{TradingEnv}.
    \item \textbf{Visual Input.}~~This characteristic is satisfied when a visual component constitutes a portion of the state space. For example, in \texttt{RoboFeederEnv}, the observation space is defined by camera images.
\end{itemize}

\textbf{RL Paradigms.}

\begin{itemize}
    \item \textbf{Frequency Adaptation.}~~Given that many real-world problems can be naturally defined in the \emph{continuous time} domain, the environments within this paradigm allow selecting actions at different frequencies. Increasing the control frequency of the system offers the agent more control opportunities, at the cost of higher computational and sample complexity. \texttt{MicrogridEnv} and \texttt{TradingEnv} are natural examples in which different market opportunities can be exploited at various frequencies.
    \item \textbf{Hierarchical RL.}~~The hierarchical RL (HRL) paradigm allows modeling, at different specificity levels, complex real-world problems presenting an inherent stratified structure. HRL divides the problem into simpler subtasks arranged in a hierarchy, thus enabling more efficient learning. Notably, \texttt{RoboFeederEnv} represents an example where the tasks of ordering (\texttt{RoboFeeder-Planning}) and collecting (\texttt{RoboFeeder-Picking}) items are hierarchically decoupled.  
    \item \textbf{Risk-Averse.}~~Risk-averse RL focuses on mitigating uncertainty by favoring policies that yield more predictable and less variable outcomes. This paradigm is especially relevant in high-stakes real-world domains, such as finance (\texttt{TradingEnv}) or healthcare, where large losses can have significant consequences.
    \item \textbf{Imitation Learning.}~~Imitation Learning in RL refers to a class of methods in which an agent learns to perform tasks by mimicking expert behavior. To this extent, a dataset comprising expert choices is needed. Within our \texttt{DamEnv} and \texttt{WDSEnv} environments, we provided such data information.
    \item \textbf{Provably Efficient.}~~Provably Efficient RL refers to a class of algorithms that come with theoretical guarantees on their performance, typically in terms of sample or computational efficiency. These guarantees can be tested more easily in tabular settings, such as the one provided in \texttt{ElevatorEnv}.
    \item \textbf{Multi-Objective RL.}~~Multi-Objective RL (MORL) is a paradigm where the agent must optimize multiple contrastive objectives simultaneously, instead of a single scalar reward. Multiple contrastive objectives can be identified in \texttt{DamEnv}, \texttt{MicrogridEnv}, and \texttt{WDSEnv}, where a trade-off between competing interests needs to be simultaneously optimized.
\end{itemize}

%% file: appendix/datasets.tex
\section{Datasets}
\subsection{\texttt{DamEnv}}

The dataset underlying the current training of the \texttt{DamEnv} environment is related to Lake Como, Northern Italy. The dataset includes historical daily records of water level, demand, and inflow from 1946 to 2010. %Figure~\ref{fig:dam_dataset_plot} reports the average data trends throughout the year. 
The 65 one-year-long time series have been split into train and test sets in an $80$\%–$20$\% proportion, resulting in 52 years for training and 13 subsequent years for testing. Additionally, the minimum required water release for each day of the year can be integrated into the dataset.
Furthermore, given the information of the lake water level, it is possible to reconstruct the control action performed by experts, therefore enabling the use of this environment to perform Imitation Learning or Inverse RL tasks. 
Data are released under a Creative Commons Attribution-NonCommercial 4.0 International (CC BY-NC 4.0) license.\footnote{\label{cc_by-nc4}\url{https://creativecommons.org/licenses/by-nc/4.0/}}

% \begin{figure}[ht]
% \centering
% \begin{minipage}{.49\textwidth}
%     \subfloat[Mean demand and inflow]{%
%   \includegraphics[width=\linewidth]{appendix/imgs/dam_demand_inflow.pdf} \label{fig:dam_demand_inflow}
% }
% \end{minipage}
% \hfill
% \begin{minipage}{.49\textwidth}
%    \subfloat[Mean lake water level]{%
%    \includegraphics[width=\linewidth]{appendix/imgs/dam_level.pdf} \label{fig:dam_level}
% } 
% \end{minipage}
% \caption{Plots of the dam dataset means for every day of the year with $95\%$ confidence intervals.}
% \label{fig:dam_dataset_plot}
% \end{figure}

\subsection{\texttt{ElevatorEnv}}
\texttt{ElevatorEnv} does not rely on external datasets. Instead, passenger arrival profiles are synthetically generated using a Poisson process, with the arrival rate for each floor independently sampled from a uniform distribution within a predefined range. This range is carefully selected to avoid unrealistic scenarios, such as excessively high arrival rates that exceed the physical handling capacity of the elevator. Specifically, the arrival rate $\lambda_f$ for each floor $f$ is sampled from the interval $[0.01, 0.1]$, resulting in a minimum and maximum expected number of arrivals of approximately $43$ and $372$, respectively, over a one-hour window with arrivals occurring at one-second intervals.

\subsection{\texttt{MicrogridEnv}}
\texttt{MicrogridEnv} integrates diverse datasets covering energy consumption, renewable energy generation, market pricing, and ambient temperature. Most datasets are provided at hourly resolution, except for ambient temperature, which is available daily. The data span the period from 2015 to 2021. We use the first four years for training and reserve the final year for testing.

\paragraph{Energy Consumption.} 
This dataset includes $398$ realistic demand profiles representing residential electricity usage across different Italian regions (North, Center, South, and islands), obtained from~\citet{fioriti2022optimal}. We split these demand time-series in $370$ profiles for training and the remaining $28$ for testing. The dataset is released under a Creative Commons Attribution 4.0 International (CC BY 4.0) license.\footnote{\label{cc_by4}\url{https://creativecommons.org/licenses/by/4.0/}}

\paragraph{Energy Generation.}
Hourly renewable generation profiles are computed for a 3\,kW residential photovoltaic installation located in northern Italy. Capacity factors are derived from~\citet{pfenninger2016longterm} and retrieved via \url{https://www.renewables.ninja}, under a CC BY-NC 4.0 license\cref{cc_by-nc4}.

\paragraph{Energy Market.}
Electricity prices are based on hourly data from the Italian Gestore Mercati Energetici (GME)~\cite{GME}, including the National Single Price and fixed tax-related and operational fees. These data, available under the CC BY-NC 4.0 license, are used to define the energy selling price. The buying price includes an additional empirical surcharge to account for taxes and provider fees.

\paragraph{Ambient Temperature.}
Daily temperature data are sourced from~\citet{staffell2023global} and obtained from \url{https://www.renewables.ninja}, also under the CC BY-NC 4.0 license\cref{cc_by-nc4}.

\subsection{\texttt{RoboFeederEnv}}
%The \texttt{robofeeder} environments are not using any external dataset.
%In the version \texttt{picking-v0} the object detection used is trained on a set of 300 synthetics images, generated from the simulator itself, and trained in an offline mode, respect to the environment dynamics.
The \texttt{RoboFeederEnv}'s tasks do not rely on any external dataset. In particular, in the \texttt{picking-v0} environment, the object detection module is trained offline using a set of $300$ synthetic images generated directly from the simulator. The geometry of the object is not synthetic or artificially designed, but it is derived directly from real-world components used in the automotive industry. Specifically, it reflects the exact structure of metallic hinges employed in assembly applications, ensuring that all evaluations are grounded in practical, industrially relevant scenarios. This training process is independent of the environment's dynamics and interaction loop, ensuring a clear separation between perception and control.

\subsection{\texttt{TradingEnv}}

TradingEnv is built using historical market data. In this suite, we consider the Forex market, in which currencies are traded. In particular, in the environment, we provide an implementation for the EUR/USD currency pair. The Forex market is open $24$ hours a day, $5$ days a week.
In this environment, we use tick-level data freely available at \url{https://www.histdata.com/}, which provides the highest level of granularity by recording every individual price change.
Each entry in the dataset contains the datetime with nano-second resolution, bid price, ask price, and volume, which is then resampled to a minute resolution. 
%(which is not populated and is always set to $0$), 

\subsection{\texttt{WDSEnv}}
\texttt{WDSEnv} relies on synthetically generated residential water demand datasets using STREaM~\cite{cominola2016developing}, a stochastic simulation model to generate synthetic time series of water end uses with diverse sampling resolutions. STREaM is calibrated on a large dataset that includes observed and disaggregated water end-uses from U.S. single-family households.  The STREaM simulator is available under the GNU General Public License 3.0 (GPL-3.0)\footnote{\url{https://www.gnu.org/licenses/gpl-3.0.html}}.

We categorized the generated profiles into one of three classes, representing increasing levels of stress imposed on the water distribution system: \textit{normal}, \textit{stressful}, or \textit{extreme}. These classes are designed to facilitate the analysis of network resilience under varying conditions.

During training, users can specify the probability distribution over the three demand types via the environment’s configuration file. Specifically, the dataset includes 21 files corresponding to \textit{normal} operating conditions, 5 files reflecting \textit{stressful} scenarios, and 5 files depicting \textit{extreme} demand situations. Each file comprises a time series of weekly demand values over a period of 53 weeks. The test set comprises $1,\!000$ synthetic profiles, held out from training and randomly sampled across the defined demand categories.

%% file: appendix/env_details.tex
\section{Environment Details}

\subsection{\texttt{DamEnv}} \label{app:dam-details}

In addition to the description in Section~\ref{sec:dam}, in this section we provide further details about the \texttt{DamEnv} environment. This environment models a water reservoir with the aim of controlling the water release to meet the demand while avoiding floods and starvation. The action is provided daily, and it represents the amount of water per second to be released during the day.

\paragraph{Simulator.}
The dynamics of the reservoir are modeled with greater granularity with respect to the action, which is provided daily. This is controlled by the parameter \texttt{integration} in the environment configuration file. It is set by default to $24$, which corresponds to one iteration per hour.

For each iteration, the bounds of the feasible amount of water to release are computed as explained in the following paragraph. Those bounds then clip the action, and the new water level is calculated. In this calculation, the water inflow and the evaporation (if provided at initialization) are also taken into account.

In Table~\ref{tab:dam_params} we report the default parameters of the simulator, which refer to the Lake Como basin, in Northern Italy. We set \texttt{evaporation} to \texttt{False}. If set to \texttt{True}, \texttt{evaporation\_rates} must be provided in the form of a csv file path with a value for each day of the year. Many of these parameters are needed for clipping the action, and their role will be clarified in the following paragraph.

\renewcommand{\arraystretch}{1.3} % Adjust row spacing
\begin{table}[ht]
    \centering
    \caption{\texttt{DamEnv}'s simulator parameters.}
    \vspace{0.1 cm}
    \rowcolors{2}{gray!10}{white} % Alternate row colors
    \begin{tabular}{l c r}
        \hline  
        \rowcolor{gray!30}
        \textbf{Parameter} & \textbf{Symbol} & \textbf{Value} \\
        \hline
        Surface & $A$ & $145.9$ [km$^2$] \\
        Evaporation & - & False [-] \\
        Initial level & $l_0$ & $0.35$ [m] \\
        Minimum flow & $r_{\text{min}}$ & $5$ [m$^3/$s] \\
        Minimum level & $l_{\text{min}}$ & $-0.5$ [m] \\
        Maximum level & $l_{\text{max}}$ & $1.25$ [m] \\
        Zero-flow level & $\alpha$ & $-2.5$ [m]\\
        Rating exponent & $\beta$ & $2.015$ [-]\\
        Discharge coefficient & $C_r$ & $33.37$ [m$^{3-\beta}$/s]\\
        Linear slope & k & $1488.1$ [m$^2/$s]\\
        Linear intercept & c & $744.05$ [m$^3/$s]\\
        Linear limit & $l_{\text{lim}}$ & $-0.4$ [m]\\
        \hline
    \end{tabular}
    \label{tab:dam_params}
\end{table}

\paragraph{Feasible Actions.}
For each iteration of the lake simulator, the action is clipped in the feasible range, which is bounded by two piecewise functions. Both the maximum and the minimum account for operational and physical constraints, such as minimum water release requirements and hydraulic limits. These bounds are functions of the water level and the minimum daily release $r_t^{min}$.

The \emph{minimum release} \( q_{\text{min}}(l, t) \) on day \( t \) is defined as:
\[
q_{\text{min}}(l, t) =
\begin{cases}
0 & l \leq l_{\min}, \\
r_{\text{min}, t} & l_{\text{min}} < l \leq l_{\text{max}}, \\
C_r (l - \alpha)^\beta & l > l_{\text{max}} \,.
\end{cases}
\]

This structure ensures that: no water is released when the water level is below the critical level \( l_{\text{min}} \), the minimum release \( r_{\text{min}, t} \) is maintained within the normal operating range, and a nonlinear release based on a rating curve applies when the level exceeds \( h_{\text{max}} \).

The \emph{maximum release} \( q_{\text{max}}(h) \) is defined as:

\[
q_{\text{max}}(l) =
\begin{cases}
0 & l \leq l_{\text{min}}, \\
k l + c & l_{\text{min}} < l \leq l_{\text{lim}}, \\
C_r (l - \alpha)^\beta & l > l_{\text{lim}}.
\end{cases}
\]

This formulation reflects operational policies where: no release occurs below the critical level \( h_{\text{min}} \), a linear release policy applies in the intermediate range, and a nonlinear rating curve governs the release when the water level exceeds \( l_{\text{lim}} \).

\paragraph{Reward Function.}
The reward function of \texttt{DamEnv} presents five components:
\begin{itemize}
    \item \textbf{Daily deficit}: it penalizes actions that do not meet the daily demand. It is computed as \(-\max \left( d_t - \max\left( r_t - r_{\text{min}, t}, 0\right), 0 \right)\), where at time $t$, $d_t$ is the demand, $r_t$ is the water that is being released, and $r_{\text{min}, t}$ is the minimum amount of water that needs to be released.
    \item \textbf{Overflow}: it penalizes actions that lead the water level to go beyond the overflow threshold. If this happens, its value is $-1$, otherwise $0$.
    \item \textbf{Starving}: similarly, it penalizes actions that lead the water level to go below the starving threshold. If this happens, its value is $-1$, otherwise $0$.
    \item \textbf{Wasted water}: it penalizes actions that release more water than the demand. It is computed as \(-\max \left( r_t - d_t, 0 \right)\), where $r_t$ and $d_t$ are defined as above.
    \item \textbf{Clipping}: it penalizes actions that do not fulfill the constraints of the environment. It is calculated as \(- \left( a_t - r_t \right)^2\) where $a_t$ is the action performed at time $t$.
\end{itemize}

Every component is weighted by a coefficient that can be set when the environment is created, balancing the influence of each component depending on the specific characteristics of the water reservoir considered.

\paragraph{Limitations.}
Simulating a different water reservoir with \texttt{DamEnv} requires domain-specific expertise to configure realistic environment parameters. Similarly, acquiring high-quality, multi-year data for other reservoirs, essential for effective learning, may be difficult.

\subsection{\texttt{ElevatorEnv}} \label{app:elevator-details}
This section provides additional details on the \texttt{ElevatorEnv} w.r.t. the main paper. The environment simulates the operation of an elevator system under a specific and realistic traffic pattern known as \textit{peak-down traffic}, which typically occurs during limited periods of the day (e.g., lunch breaks or end-of-work shifts), when the majority of users exit upper floors toward the ground level.

We adopt an episodic setting lasting $3600$ seconds (i.e., one hour), with control actions executed every second. While the environment does not include a full physical simulator, we deliberately model it in a fully discrete manner. This choice enables a compact formalization of the dynamics while preserving the core challenges of the task and allows us to use tabular policies, which are both simple and interpretable. This is particularly advantageous for benchmarking, as it avoids the complexity of large-scale function approximation while still offering meaningful learning dynamics. By carefully designing the observation space and controlling its dimensionality, we ensure that the problem remains tractable, yet rich enough to pose a non-trivial challenge. Additionally, considering the six \texttt{Gym4ReaL} environments, this is the only one that allows users to test RL algorithms designed for tabular settings, as typical of provably efficient approaches, thus broadening the applicability of our library.

The configurable parameters used in this task are reported in Table~\ref{tab:elevator_params}.

\renewcommand{\arraystretch}{1.3} % Adjust row spacing
\begin{table}[th]
    \centering
    \caption{\texttt{ElevatorEnv} parameters.}
    \vspace{0.1 cm}
    \rowcolors{2}{gray!10}{white} % Alternate row colors
    \begin{tabular}{l c r}
        \hline
        \rowcolor{gray!30}
        \textbf{Parameter} & \textbf{Symbol} & \textbf{Value} \\
        \hline
        Floors & f & $\{0, \dots, 4\}$ [-] \\
        Maximum capacity & $C_{\text{max}}$ & $4$ [people]\\
        Movement speed & $v$ & $3$ [m/s] \\
        Floor height & - & $6$ [m] \\
        Maximum queue length & $W_{f, \text{max}}$ & $3$ [people]\\
        Maximum new arrivals & - & $2$ [people]\\
        Arrival rate & $\lambda_f$ & $[0.01, 0.1]$ [-]\\
        \hline
    \end{tabular}
    \label{tab:elevator_params}
\end{table}

\paragraph{Assumptions.} 
The following assumptions are made to simplify the dynamics of the \texttt{ElevatorEnv}, allowing for an efficient yet representative modeling of the task:
\begin{itemize}
    \item \textbf{No acceleration dynamics.} We do not model acceleration or deceleration; the elevator is assumed to move with uniform rectilinear motion;
    \item \textbf{Instantaneous boarding.} We assume that the entry and exit of passengers inside the elevator occur instantaneously;
    \item \textbf{Uniform floor spacing.} All floors are equidistant, and the total height of the building is assumed to be divisible by the elevator's speed. This ensures that the elevator can only occupy a finite number of discrete vertical positions;
    \item \textbf{Elevator height.} For the same reason as the previous bullet point, we assume that the height of the elevator is equal to the floor height;
    \item \textbf{Non-floor stopping.} We allow the elevator to open its doors even between two floors, unlike in real-world systems. While this behavior is physically unrealistic, it does not provide any advantage. The agent is penalized for unnecessary stops and will learn to avoid such actions through experience.
\end{itemize}

\paragraph{Feasible Actions.} 
We adopt a permissive design where all control actions are considered feasible at any state. In particular, we do not explicitly penalize the agent for executing physically invalid actions, such as opening doors between two floors (e.g., when the floor height is smaller than the elevator's movement speed), or attempting to move upward at the top floor ($a_t = \textit{u}$ when $h_t = H$), or downward at the ground floor ($a_t = \textit{d}$ when $h_t = 0$). These actions do not result in termination or constraint violations but are treated as valid no-ops since the agent is still implicitly penalized through time progression, as ineffective or wasteful actions delay task completion and reduce the cumulative reward.

\paragraph{Reward Function.}
Although not strictly necessary to solve the task, the $\beta$ hyperparameter plays a crucial role in accelerating the learning process. Specifically, $\beta$ modulates the only source of positive reward available to the agent, i.e., passenger delivery. This design encourages the agent to recognize that minimizing the cumulative waiting time inherently requires completing passenger trips by bringing them to their destinations. In doing so, $\beta$ reinforces the primary operational goal of an elevator system: transporting users efficiently rather than merely avoiding penalties or idle behavior.

\paragraph{Limitations.}
The principal limitation of \texttt{ElevatorEnv} arises from its scalability. While the environment is inherently finite and conceptually straightforward, the dimensionality of its observation space can increase rapidly, rendering it intractable for tabular RL algorithms. Consequently, it is essential to carefully configure the environment, adjusting its parameters to ensure that the resulting problem is neither trivial nor computationally prohibitive.

\subsection{\texttt{MicrogridEnv}} \label{app:microgrid-details}
This section provides further details on \texttt{MicrogridEnv}. As outlined in the main paper, this environment models a control problem centered on optimal electrical energy management in a residential setting equipped with a PV implant, a battery system, and access to the main grid for energy trading.
At each time step $t$, the controller must decide how to allocate the net power, $P_{N,t} = P_{G,t} - P_{D,t}$ where $P_{G,t}$ is the power generated by the PV system and $P_{D,t}$ is the residential electricity demand. The agent must choose whether to address the net power to the battery or trade it with the grid. Importantly, this decision is made based on estimates of $P_{G,t}$ and $P_{D,t}$, as the actual values are unknown at decision time.

Consequently, the decision-making process is inherently influenced by uncertainty in both demand and generation forecasts, as well as by exogenous factors such as electricity prices and ambient temperature, each exhibiting cyclo-stationary patterns. Notably, ambient temperature significantly impacts battery degradation, creating a long-term trade-off between maximizing short-term energy efficiency and preserving battery health.

The problem is formulated as an infinite-horizon task, which terminates when the battery reaches its end-of-life (EOL) condition. Control actions are taken every $3600$ seconds, matching the finest available data granularity.

\paragraph{Simulator.} 
We simulate the dynamics of the BESS using a digital twin~\cite{salaorni2023ernesto}, which models the evolution of the battery's state of charge (SoC), temperature, and state of health (SoH) over time. The simulator emulates a realistic lithium-ion battery pack suitable for residential applications.
The simulator operates in a step-wise manner consistent with the Gymnasium interface. At each time step, it receives the input power to store or retrieve from the battery, computes the resulting voltage, thermal dynamics, SoC update, and degradation, and returns a snapshot of the battery's status to the RL agent.

Model parameters and configurations are based on expert consultation and are preconfigured for direct use. However, users can easily modify or replace the battery model configuration to experiment with different setups or refer directly to the full simulator details in~\cite{salaorni2023ernesto}. The specific parameters used in this work are summarized in Table~\ref{tab:mg_params}.

\renewcommand{\arraystretch}{1.3} % Adjust row spacing
\begin{table}[th]
    \centering
    \caption{\texttt{MicrogridEnv}'s simulator parameters.}
    \vspace{0.1 cm}
    \rowcolors{2}{gray!10}{white} % Alternate row colors
    \begin{tabular}{l c r}
        \hline
        \rowcolor{gray!30}
        \textbf{Parameter} &  \textbf{Symbol} & \textbf{Value} \\
        \hline
        Nominal capacity & $C_N$ &$60.0$ [Ah] \\
        Maximum voltage   & $V_{\text{max}}$ & $ 398.4$ [V] \\
        Minimum voltage   & $V_{\text{min}}$ & $ 288.0$ [V] \\
        Replacement cost  & $\mathcal{R}$ & $3000$ [€] \\
        SoC range   & - & $[0.2, 1]$ [-] \\
        \hline
    \end{tabular}
    \label{tab:mg_params}
\end{table}

\paragraph{Observation Space.}
The observation space includes the variables $\widehat{P}_{D,t}$, $\widehat{P}_{G,t}$, $p^{buy}_t$, and $p^{sell}_t$, which encapsulate the exogenous factors influencing the system, namely, energy demand, renewable energy generation, and the buying/selling energy prices. Ideally, the control decision at time $t$ should rely on the actual values of demand $P_{D,t}$ and generation $P_{G,t}$. However, these values are not available at decision time. To overcome this, we approximate them using their most recent observations at time $t-1$, denoted as $\widehat{P}_{D,t}$ and $\widehat{P}_{G,t}$. This assumption is reasonable in typical microgrid scenarios, where energy demand and generation profiles exhibit temporal smoothness or autocorrelation, allowing previous values to serve as informative estimators. On the other hand, energy market prices $p^{buy}_t$ and $p^{sell}_t$ are assumed to be known in advance. This assumption is justified by the common practice in energy markets where prices are typically set a day ahead, enabling the agent to access this information at decision time.

To further enhance the expressiveness of the observation space, we introduce two time-based features that encode temporal periodicity. These are defined as:
\begin{itemize}
    \item $\varphi^d = \frac{2\pi \tau_d}{T_d}$, where $\tau_d \in [0, T_d]$ represents the current time of day in seconds, and $T_d$ is the total number of seconds in a day. This captures daily periodicity (e.g., day-night cycles).
    \item $\varphi^y = \frac{2\pi \tau_y}{T_y}$, where $\tau_y \in [0, T_y]$ represents the current time of year in seconds, and $T_y$ is the total number of seconds in a year. This accounts for seasonal patterns within datasets.
\end{itemize}

\paragraph{Feasible Actions.}
The controller's action $a_t$ defines the fraction of the net power $P_{N,t}$ to allocate to the battery at time step $t$. We can formally express this decision as:
\begin{align}
    P_{B,t} & = a_t P_{N,t}, \\
    P_{E,t} & = (1 - a_t) P_{N,t},
\end{align}
where $P_{B,t}$ is the amount of power stored in (or retrieved from) the battery, and $P_{E,t}$ is the portion traded with the main grid, either sold or purchased.

The controller's behavior depends on the sign of $P_{N,t}$:
\begin{itemize}
    \item \textbf{Deficit case ($P_{N,t} < 0$)}: The generated PV power does not meet the demand. The controller determines $P_{B,t}$, the amount to draw from the battery, and covers the remaining deficit $P_{E,t}$ by purchasing from the grid.
    \item \textbf{Surplus case ($P_{N,t} > 0$)}: There is excess generation relative to demand. The controller allocates $P_{B,t}$ to the battery and sells the remaining $P_{E,t}$ to the grid.
\end{itemize}

The chosen action must respect the physical constraints of the BESS, such as bounds on charging/discharging power and SoC limits. At time $t$, SoC denoted as $\sigma_t \in [0,1]$ is defined by:
\begin{equation}
    \sigma_t := \sigma_{1} + \sum_{h=2}^t \frac{i_h \Delta \tau}{C_h},
\end{equation}
where $\sigma_1$ is the initial SoC, $i_h$ is the current, $C_h$ is the internal battery capacity at time $h$, and $\Delta \tau$ is the time step in hours. 
To ensure valid operations, $P_{B,t}$ must satisfy:
\begin{align}
    P_{dch} &\leq P_{B,t} \leq P_{ch}, \label{eq:constr1} \\
    \frac{(\sigma_{\min} - \sigma_t)}{\Delta \tau} C_t V_t &\leq P_{B,t} \leq \frac{(\sigma_{\max} - \sigma_t)}{\Delta \tau} C_t V_t, \label{eq:constr2}
\end{align}
where $P_{dch} < 0$ and $P_{ch} > 0$ are the maximum discharging and charging powers, $\sigma_{\min}$ and $\sigma_{\max}$ are the minimum and maximum SoC levels, and $V_t$ is the battery voltage at time $t$.

\paragraph{Reward Function.}
The reward signal consists of three components: trading profit, battery degradation cost, and a penalty for violating physical constraints. Formally, the trading component $r_{\text{trad}}(a_t)$ of the reward is defined as:
\begin{align}
    r_{\text{trad}}(a_t) = \left(p_t^{\text{sell}} P^+_{E,t} + p_t^{\text{buy}} P^-_{E,t}\right) \Delta \tau \,, \label{eq: r_trad}
\end{align}
where $p^{\text{sell}}_{t}$ and $p^{\text{buy}}_t$ are the unit energy prices for selling and buying electricity at time $t$, respectively, with $p^{\text{sell}}_{t} < p^{\text{buy}}_t$. The terms $P^+_{E,t}$ and $P^-_{E,t}$ denote the positive and negative parts of $P_{E,t}$, respectively.\footnote{For a real-valued quantity $q \in \mathbb{R}$, its positive and negative parts are defined as $q^+ := \max\{0, q \}$ and $q^- := \min \{0, q \}$, respectively.}

The degradation component $r_{\text{deg}}(a_t)$ accounts for battery aging and is linked to the SoH, which monotonically decreases from $100\%$ to an application-specific EOL threshold (typically $60$--$80\%$). Let $\rho_t \in [0,1]$ denote the SoH at time $t$, defined as:
\begin{equation}
    \rho_t := \frac{C_t}{C_N},
\end{equation}
where $C_t$ is the internal capacity at time $t$, and $C_N \in \mathbb{R}^+$ is the nominal capacity of the system. The SoH decay incorporates both calendar aging and usage-related degradation. Thus, the degradation cost is computed as:
\begin{equation}
    r_{\text{deg}}(a_t) = \frac{\rho_t - \rho_{t-1}}{1 - \rho_{\text{EOL}}} \mathcal{R}, \label{eq: r_deg}
\end{equation}
where $\rho_{\text{EOL}} \in (0, 1)$ is the SoH value at end-of-life, and $\mathcal{R}$ is the BESS replacement cost. Note that since $\rho_t < \rho_{t-1}$, this term yields a negative reward.

Finally, $r_{\text{clip}}(a_t)$ penalizes actions that violate the system's operational constraints and must be clipped. In our formulation, we only penalize violations of the tighter constraint~\eqref{eq:constr2}, which is stricter than~\eqref{eq:constr1}. Hence, formally, the penalty is defined as:
\begin{equation}
    r_{\text{clip}}(a_t) = - \max \Big\{ 0, \, a_t P_{N,t} + \frac{(\sigma_t - \sigma_{\max})}{\Delta \tau} C_t V_t, \frac{(\sigma_{\min} - \sigma_t)}{\Delta \tau} C_t V_t - a_t P_{N,t} \Big\}.
\end{equation}

\paragraph{Limitations.}
The main limitation of \texttt{MicrogridEnv} lies in its data and configuration specificity. The environment is built upon data representative of a single geographical and regulatory context (Italy), which includes particular consumption patterns, solar generation profiles, and market prices. Additionally, the battery configuration models a specific lithium-ion storage system with fixed chemical composition, physical dimensions, and operational parameters. As a result, the environment may not generalize directly to other geographical regions or technological setups without additional data integration or customization.

\subsection{\texttt{RoboFeederEnv}} \label{app:robofeeder-details}
\begin{figure}[ht!]
    \centering
        % Second image
    \begin{minipage}[b]{0.42\textwidth}
        \centering
        \includegraphics[width=\linewidth, trim={4.5cm 0 4.5cm 0}, clip]{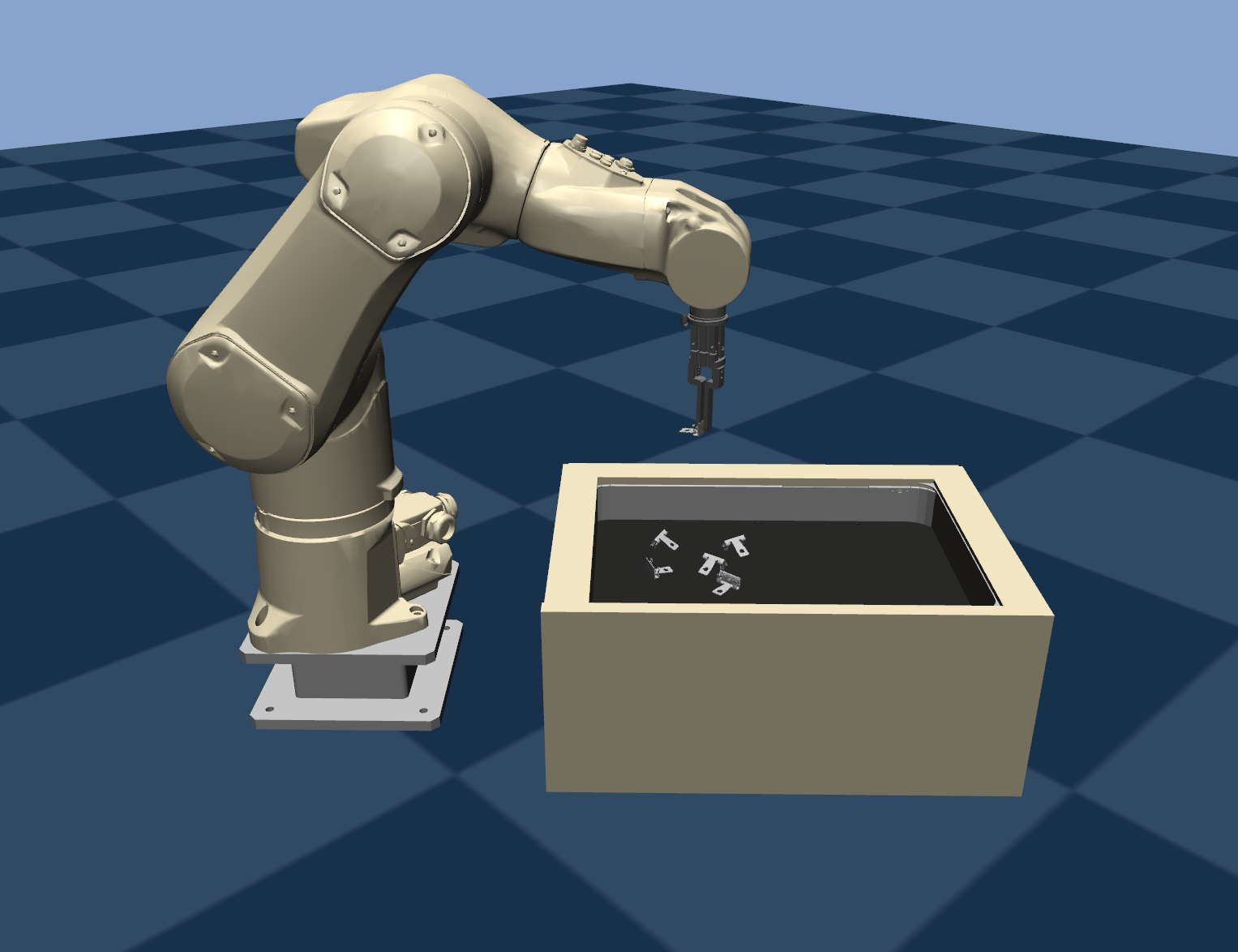}
        \caption{Rendering of MuJoCo simulator from \texttt{RoboFeederEnv}.}
        \label{fig:robofeeder_sideview}
    \end{minipage}
    \hfill
    \begin{minipage}[b]{0.42\textwidth}
        \centering
        \includegraphics[width=\linewidth]{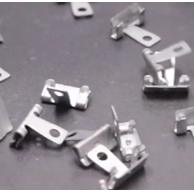}
        \caption{ the objects considered, metallic hinges from the automotive sector (\cite{depaola2024power}).}
        \label{fig:robofeeder_obj}
    \end{minipage}
\end{figure}

In addition to the description provided in Section~\ref{sec:robofeeder}, we present further details used in designing the \texttt{Robofeeder} environment. \texttt{Robofeeder} (Fig.~\ref{fig:robofeeder_sideview}) is designed to handle small objects like those shown in Figure~\ref{fig:robofeeder_obj}, using a commercial 6-DOF Staubli robot with a predefined kinematic cycle that enables reliable pick-and-place operations. The environment includes a configuration file that allows users to specify the number of objects in the scene and their initial orientations. Notably, since \texttt{RoboFeederEnv} is built on MuJoCo, the object geometries can be easily modified, making it possible to extend the simulator's capabilities far beyond the examples presented here.
In Table~\ref{tab:rf_hyperparams}  we report the main parameters of the \texttt{RoboFeederEnv} environments.

\renewcommand{\arraystretch}{1.3} % Adjust row spacing
\begin{table}[th]
    \centering
    \rowcolors{2}{gray!10}{white} % Alternate row colors
    \caption{\texttt{RoboFeeder} 
    Parameters.}
    \vspace{0.1cm}
    \resizebox{\textwidth}{!}{
    \begin{tabular}{lcrrr}
        \hline
        \rowcolor{gray!30}
        \textbf{Parameter} & \textbf{Symbol} & \texttt{picking-v0} & \texttt{picking-v1} & \texttt{planning} \\
        \hline
        \text{\# objects} & N & $1$ & $1$ & $3$\\ 
        \text{Observation shape} & - & $1\times 50 \times 50 \times 1$ [px] & $1\times300\times300\times1$ [px] & $3\times50\times50\times1$ [px]\\ 
        \text{Distance-rotation weight} & \text{$\lambda$} & $0.85$ & $0.85$ & - \\ 
        \text{Threshold distance} & \text{$\delta_{d}$} & $0.012$ & $0.012$ & -\\ 
        \text{Threshold rotation} & \text{$\delta_{\theta}$} & $0.35$ & $0.35$ & -\\ 
        \text{Distance decay rate} & \text{$\alpha_d$} & $-\frac{\ln(0.5)}{(\delta_d)^2}$ & $-\frac{\ln(0.5)}{(\delta_d)^2}$  & -\\
        \text{Rotation decay rate } & \text{$\alpha_\theta$} & $-\frac{\ln(0.5)}{(\delta_\theta)^2}$ & $-\frac{\ln(0.5)}{(\delta_\theta)^2}$ & -\\
        \text{Timestamps} & \text{$T$} & $2$ & $2$ & $5$\\
        \hline
    \end{tabular}
    }
    \label{tab:rf_hyperparams}
\end{table}

\subsubsection{\texttt{RoboFeeder-picking}}
In each episode, the agent interacts with the family of \texttt{RoboFeeder-picking} environments by selecting an action $a_0$ based on the initial state $s_0$ of the system, where a single object is placed within the robot's working area. The action $a_0 \in [-1, 1]$ is rescaled with respect to the observation space to map it onto the $u$-$v$ image coordinates, representing a precise position in the MuJoCo robotic world. After executing the action, the agent receives a reward $r(s_0, a_0)$, and the simulator transitions to state $s_1$ depending on the success of the robot's motion. This process repeats at each timestep until the episode reaches the terminal step $t = T$.

\paragraph{Assumptions.}
The picking environments are designed to train a specialized robot to grasp a single, small, and complex-geometry object. To ensure the robot can consistently focus on the picking task, the environment contains only one object, which is placed in a way that always allows it to be successfully grasped. Even though the object's position is randomized at the start of each episode, its orientation is carefully set to ensure that at least one feasible robot pose exists from which it can be picked. When working with 6-DOF robots, mechanical constraints may prevent the end-effector from reaching every position in Cartesian space. Therefore, robot programming should avoid assigning target poses that are near kinematic singularities or that would place excessive stress on the joints. To replicate reality, we exclude from the feasible action space the upper part of the workspace area, not reachable by the robot. This assumption could be changed from the base configuration file provided in the environment implementation.

\paragraph{Observation Space.}
\texttt{Gym4Real} implements two versions of this environment. \texttt{picking-v0} represents the simpler of the two, as the region of interest in which the agent must search is limited to the object's neighborhood, thanks to the assistance of a pretrained object detection neural network. By leveraging an SSD MobileNet V2 network~\cite{mobilenet}, the observation space is significantly reduced to a cropped image of dimensions $1 \times H_{\text{cropped}} \times W_{\text{cropped}} \times C$, where $H_{\text{cropped}} = W_{\text{cropped}} = 50$\,px, and $C=1$. This reduction simplifies the task, as the reward function is designed to evaluate the distance between the robot's selected grasp point and the actual object. By narrowing the search area, the agent can focus on a more relevant region of the scene, reducing training time and avoiding sparse experience with little to no learning signal.

\paragraph{Feasible Actions.}
Actions coincide with a continuous value that normalizes the choice of where to grasp, depending on the observation space definition. It means that based on the observation, i.e., the actual visual input, the agent selects the portion of the image to identify as a grasping point. From the image projection, the environment converts the agent's choice to a unique robot world position, which is relative to the fixed position of the top-down camera in the world. The latter is set parallel to the object's \textit{home} position, acquiring a fixed portion of the scene. This design allows for playing an action with a granularity definition in the continuous domain.

\paragraph{Reward Function.}
The reward function for the \texttt{picking} environments is defined as a weighted sum of the distance and orientation alignment between the robot's end-effector (TCP) and the target object. %Exception to this happens when the object is placed in the final expected position.

Let \( d \) denote the Euclidean distance between the TCP and the object, and \( \Delta\theta \) the angular misalignment (in radians) between the TCP and the object. Define \( \delta_{\text{lim}}\) as the distance threshold for soft negative feedback, and \( \delta_{\theta}\) as the angular threshold for rotational alignment. A weighting parameter \( \lambda \in [0, 1] \) balances the contribution of the distance and rotational terms.

The reward function \( r(d, \Delta\theta) \) is then given by:

\begin{equation}\label{eq:reward-picking}
    r_t(d, \Delta\theta) = 
    \begin{cases}
    1 - \left[\lambda \cdot \exp\left(\alpha_{\text{d}} \cdot d^2\right) + (1-\lambda) \cdot \exp\left(\alpha_{\theta} \cdot (\Delta\theta)^2\right) \right] &  \text{picking failed}\\
    1 & \text{picking successfull}
\end{cases}
\end{equation}

$\alpha_d$ and $\alpha_\theta$ act as decay rates to incentivize the agents to correctly pick the object, where $\alpha_d = -\frac{\ln(0.5)}{(\delta_d)^2} $ and 
$\alpha_\theta =  -\frac{\ln(0.5)}{(\delta_\theta)^2}$.

This formulation ensures that when \( d = \delta_{\text{d}} \) or \( \Delta\theta = \delta_{\theta} \), the respective exponential term evaluates to \( 0.5 \). The parameter \( \lambda \) thus tunes the agent's sensitivity between spatial precision and angular alignment during grasping attempts.
Both versions of the picking environments share the parameter $\lambda$, which expresses the preference of the simulator for giving a higher reward when correctly approaching the theoretical point of grasp in the Euclidean space with respect to its correct orientation. $\lambda$ is set to encourage the agent to reach a limit distance in terms of space and angle, so that it will be able to correctly pick the object.
The cumulative reward is updated as $\sum_t^T r_t(d,\Delta_\theta)$.

\paragraph{Limitations.}
The \texttt{picking} environments are well-suited for scenarios where the user needs to adjust the object geometry. However, the embedded kinematic solver is specifically designed for the robot being used, which means it does not support modifications to the robot model. This limitation extends to the two-finger grasping tool and its pick-and-place cycle.

\paragraph{Why RL is Needed.}
Although the \texttt{RoboFeeder-picking} problem may initially appear solvable through supervised learning, given that the object is always placed in a pickable position, the task fundamentally requires RL. This is due to the need for the agent to learn grasping as a procedural strategy, rather than a simple input-output mapping. In practice, it is not sufficient to identify where the object is located; the agent must also learn how to approach and align itself to perform a successful grasp under realistic constraints, such as occlusions, partial views, and kinematic limitations of the robot. While in a real-world setting, a human expert could manually specify the grasp points, the objective of this environment is to enable the robot to autonomously learn the entire grasping process—from perception to actuation—without supervision. This involves discovering optimal action sequences and dealing with delayed rewards and sparse success signals, challenges that are inherently better addressed through RL frameworks rather than supervised data-driven approaches.

\subsubsection{\texttt{RoboFeeder-planning}}
In each episode, the agent interacts with the \texttt{RoboFeeder-planning} environment by selecting a discrete action $a_0$ based on the initial state $s_0$, where multiple objects are placed within the robot's working area. The action $a_0 \in \{0, \ldots, n\}$, with $n \in \mathbb{N}$, corresponds to a relative index in a vector of cropped images representing the state $s_0$. From $s_0$, the selected cropped image, of shape $1 \times H_{\text{cropped}} \times W_{\text{cropped}} \times C$, is used as input to a pre-trained agent that determines a grasp point to reach an object, if one is present in the selected region.
Depending on the success of the grasp attempt, the agent receives a reward $r(s_0, a_0)$, and the simulator transitions to a new state $s_1$. This process is repeated at each time step $t \in [T] \coloneqq \{1, \dots, T\}$ until the episode reaches the final step $t = T$.

\paragraph{Observation Space.}
The environment is designed to handle a variable number of objects within the robot's workspace. This is achieved by maintaining a fixed-size array of detected objects. The visual observation at each timestep is represented as
$s_t = [\mathbf{X}_{1,t}, \dots, \mathbf{X}_{N,t}], \quad \mathbf{X}_{i,t} \in \mathbb{R}^{H \times W \times C}$
where each \( \mathbf{X}_{i,t} \) corresponds to the image crop of the \(i\)-th detected object. Although the environment can potentially detect many objects, we define an upper bound \( N = 3 \) on the number of objects processed in each observation, ensuring a consistent and manageable input size for the agent.
Each cropped image, used in observation space, contains the result of the object detection stage, capturing only one object at a time and creating a totally black image when the number of objects present in the robot workspace is less than $N$.
This would happen if the environment is loaded with fewer objects or during the episode, simulating with the robot capable of placing the objects in the final position.

\paragraph{Feasible Actions.}
Based on the construction of the observation space, the agent can only select an index $i \in [1, N]$ with $N$ the maximum fixed number of elements. object processed, to attempt a demanding grasping of the inner picking agent. The action $a_t = 0$ corresponds to a reset request.

\paragraph{Limitations.}
This environment strongly depends on the performance of a low-level pre-trained agent capable of solving the underlying \texttt{picking} tasks. The optimal policy $\pi^*(a, s)$ is one that can correctly infer when an object is properly oriented for picking, or when a sequence of grasp attempts will eventually allow the robot to feed out the remaining objects.

\subsection{\texttt{TradingEnv}}
\label{sec:env_details:tradingenv}

In this section, we provide additional details about \texttt{TradingEnv} introduced in Section~\ref{sec:tradingenv}.
The environment, for each episode, simulates a trading day, and at every step, the agent decides which action to perform to maximize the profit-and-loss (P\&L) net of transaction costs. The configuration parameters considered in the presented setting are reported in Table~\ref{tab:tradingenv_params}.

%Financial markets are known to be highly non-stationary, frequently transitioning to different market regimes, and the data related to financial securities are affected by a low signal-to-noise ratio. Those two main characteristics makes very hard to learn profitable strategies.
%Regarding the Non-Stationarity, it is important to consider the prices as price variation and/or percentage change. In addition to prices, including temporal features (e.g., minute of the day, day, month), helps the identification of temporal-based pattern. 
\paragraph{Assumptions.}
As stated in the main paper, we restricted the trading hours from 8:00 EST to 18:00 EST, from Monday to Friday, due to low liquidity outside this interval.
Hence, since we are learning an intra-day trading strategy, the position of the agent is opened at 8:00 EST and forced to be closed at 18:00 EST.

\paragraph{Observation Space.}
In this section, we want to give more information about the considered observation space. 
%the observation space is formalized as
% \begin{equation}
%     s_t = \big( \mathbf{d}_t, \text{cos}(\varphi^{day}_t), \text{sin}(\varphi^{day}_t), z_t \big),
% \end{equation}
%where $\mathbf{d}_t = [d_{0,t}, \dots, d_{59,t}]$ is the vector of the last $60$ delta mid prices at time $t$ (the delta price is formulated as $d_{k,t} = \frac{p_{t-k} - p_{t-k-1}}{p_{t-k-1}}$), $\varphi^{day}_t \in [0, 2\pi]$ is the angular position of the current time over the trading period, and $z_t=a_{t-1}$ is the agent position.
As said, the variable $\mathbf{d}_t$ is the vector of the delta mid-prices. The mid-price $p_t$ is defined as the average price between bid and ask $p_t = \frac{p_{bid} + p_{ask}}{2}$.

Furthermore, the observation space can be customized, including more and different market information, such as:
\begin{itemize}
    \item \textbf{Number of delta mid-prices:} Setting a higher number of delta mid-prices could allow for considering longer price evolution. However, this could increase noise in the observations, negatively impacting the learning.
    \item \textbf{Delta mid-prices offset:} Default is 1 minute, but it is possible to enlarge it to consider the return obtained with different intervals (e.g., by considering 60 delta mid-prices with 1-minute offset, we can observe information on the last hour price variation).
    \item \textbf{Temporal information:} By default, only the timestamp is considered, but it is possible to consider also Month, Day, and Day of the Week to capture more complex temporal patterns.
\end{itemize}

\renewcommand{\arraystretch}{1.3} % Adjust row spacing
\begin{table}[th]
    \centering
    \caption{\texttt{TradingEnv} parameters.}
    \vspace{0.1 cm}
    \rowcolors{2}{gray!10}{white} % Alternate row colors
    \begin{tabular}{l c r}
        \hline
        \rowcolor{gray!30}
        \textbf{Parameter} &  \textbf{Symbol} & \textbf{Value} \\
        \hline
        Number of deltas & $\mathbf{d}_t$ & $60$ [-]\\
        Offset & - & $1$ [min]\\
        Persistence & - & $5$ [min]\\
        Capital   & C & $100$k [€]\\
        Fees & $ \lambda $ &  $1$ [€] \\
        \hline
    \end{tabular}
    \label{tab:tradingenv_params}
\end{table}

Regarding the temporal features, it is also possible to enable the cyclic encoding, using sine and cosine transformations. Cyclic encoding emphasizes that time-related quantities, although numerically distant, can be close in a cyclical sense. 
%For example, 23:00 and 00:00 are numerically far but temporally adjacent. 
This type of encoding is typically preferred when using function approximators such as neural networks. In contrast, it is not recommended in some cases, such as with tree-based models, which are sometimes considered in trading settings, where the encoding can obscure the natural ordering of features.

\paragraph{Feasible Actions.}
At each decision step, the agent can perform 3 actions:
% \begin{itemize}
%     \item \textbf{}{Long}: Buy, or continue holding a fixed amount of the assets if already in long positions.
%     \item \textbf{}{Flat}: No exposition or closing position.
%     \item \textbf{}{Short}: Short-sell or continue holding a fixed amount of assets.
% \end{itemize}
As explained in Section~\ref{sec:tradingenv}, at each decision step, the agent can perform 3 actions: \textit{long}, \textit{short}, and \textit{flat}.
Each action involves a fixed amount of capital $C$ that in the experiments we set to €$100$k.

In this environment, it is possible to specify the action frequency (or \textit{persistence}), as time is discretized. The default frequency is 5 minutes, meaning that the agent reassesses its position every 5 minutes. It is important to emphasize that the persistence should be adjusted in conjunction with the observed delta mid-prices, both in terms of their quantity and the temporal offset used.

\paragraph{Reward Function.}
% The reward function is defined as:
% \begin{equation}
%     r_{t} = a_{t-1}(p_{t} - p_{t-1}) - \lambda|a_t - z_t|,
% \end{equation}
% where the first part is related to the profit-and-loss (P\&L) obtained from a specific action performed in the previous step $a_{t-1}$ and the second term is the transaction cost 
In the presented reward formulation, $\lambda$ is the transaction fees that in our experiments is set to $1$\$, which is paid for every position change. Modification to the fees parameter could lead to more conservative or aggressive strategies. 

\paragraph{Limitations.}

The main limitation of \texttt{TradingEnv} is that, being built using historical data, the effective quality of the learning environment is highly linked to the quality of those. Such data typically contains missing values that must be appropriately handled and preprocessed.
In our case, missing values up to 5 minutes (this is a customizable threshold) are forward-filled. Instead, all the days with wider gaps are removed from the data. In addition, the market impact is not taken into account. This is not a substantial limitation in high liquid settings, as the EUR/USD currency pair, considered in this paper.

We refer to Optimal Execution approaches for further details~\citep{Almgren2000OptimalEO}, since this is a well-established research area that aims to minimize market impact and prevent significant price movement during trade execution.

\subsection{\texttt{WDSEnv}} \label{app:wds-details}
\texttt{WDSEnv} addresses the problem of enhancing the resilience of residential water distribution networks. As discussed in the main paper, we rely on the EPANET simulator~\cite{epanet2000} to model the evolution of the network and to compute the fluid dynamics within the system. The task is formulated as an infinite-horizon decision-making problem, with each episode truncated after one week of simulated time. Both the hydraulic step and the demand pattern step are set to one hour, aligning with the temporal resolution typically used in real-world water management systems.

\paragraph{Simulator.}
To simulate the environment, we rely on \texttt{Epynet}~\cite{epynet}, a Python wrapper for EPANET, upon which we built our Gymnasium-based environment. While the most widely adopted Python interface for EPANET is currently \texttt{WNTR}~\cite{wntr}, it lacks support for step-wise (i.e., discrete-time) simulation of the hydraulic network~\cite{murillo2023high-fidelityI}, which is essential for reinforcement learning tasks. In contrast, \texttt{Epynet} includes built-in features that facilitate step-wise simulation with only minimal modifications required on our end.

The simulator models the water distribution system as a graph composed of nodes and links. Among the nodes, tanks, and junctions play a critical role: tanks store water and help meet demand during peak usage or scarcity and must maintain their water levels within operational boundaries; junctions, on the other hand, are expected to maintain adequate pressure to ensure service reliability.: tanks store water and help meet demand during peak usage or scarcity, and must maintain their water levels within operational boundaries; junctions, on the other hand, are expected to maintain adequate pressure to ensure service reliability. Each junction is assigned a base demand value in the \texttt{.inp} file, which represents the nominal water demand at that location. The \emph{Demand Satisfaction Ratio} (DSR) is used as a key indicator of how well the actual demand is met over time.

It is important to note that the demand values in \texttt{.inp} files are static. To simulate dynamic and diverse operational conditions, we introduce time-varying demand profiles that modulate the base demand values to reflect \textit{normal}, \textit{stressful}, and \textit{extreme} scenarios. These profiles allow us to simulate periods of increased stress on the network, such as high usage due to heatwaves or population surges. In Table~\ref{tab:wds_params}, we summarize the main parameters used to configure the water distribution simulator.

The network chosen for the experiment is the \textit{Anytown} system, which has the following composition: 22 junctions (from \textit{J}$_1$ to \textit{J}$_22$), 2 tanks (\textit{T41} and \textit{T42}), 2 pumps (\textit{P78} and \textit{P79}), and 1 reservoir (\textit{R40}).
Further details on the network configuration are available in the relative \textit{.inp} file.

\renewcommand{\arraystretch}{1.3} % Adjust row spacing
\begin{table}[th]
    \centering
    \caption{\texttt{WDSEnv}'s simulator parameters.}
    \vspace{0.1 cm}
    \rowcolors{2}{gray!10}{white} % Alternate row colors
    \begin{tabular}{l c r}
        \hline
        \rowcolor{gray!30}
        \textbf{Parameter} &  \textbf{Symbol} & \textbf{Value} \\
        \hline
        Town & - & \textit{anytown.inp} \\
        Demand estimation & $\widehat{d_t}$ & SMA\\
        Normal demand prob.   & - & $0.6$ [-]\\
        Stressful demand prob.   & - & $0.35$ [-]\\
        Extreme demand prob.   & - & $0.05$ [-]\\
        Overflow risk & $\lambda_{\text{of}}$ & $0.9$ [-] \\
        \hline
    \end{tabular}
    \label{tab:wds_params}
\end{table}

\paragraph{Observation Space.}

The observation space considered in the reported experiment includes both the tank level ($L=2$) and all the junctions of the network ($J=22$). The estimated total demand pattern $\widehat{d_t}$ is computed as a moving average over the previous quarter of the day, corresponding to a $6$-hour window. This can be implemented either as a simple moving average (SMA) or using an exponentially weighted moving average (EWMA), depending on the desired responsiveness to recent variations. 
Additionally, the observation space includes a time-based feature to capture intra-day periodicity. Specifically, we define $\varphi^d = \frac{2\pi \tau_d}{T_d}$, where $\tau_d \in [0, T_d]$ denotes the current time of day in seconds, and $T_d$ is the total number of seconds in a day (i.e., $T_d = 86400$).

\paragraph{Reward Function.} 
The reward function consists of two main components: the DSR term and the overflow penalty term. These components jointly encourage the agent to maintain an adequate water supply across the network while avoiding risky conditions such as tank overflows.

The first component, the DSR term $r_{\text{DSR},t}(a_t)$, quantifies how effectively the agent satisfies the expected demand across all junctions at time $t$. It is defined as:
\begin{equation}
    r_{\text{DSR},t}(a_t) = \frac{\sum_{j=1}^J d_{j,t}}{\sum_{j=1}^J \overline{d_{j,t}}},
\end{equation}
where $d_{j,t}$ denotes the actual supplied demand at junction $j$ and time $t$, and $\overline{d_{j,t}}$ is the corresponding expected demand. This term rewards the agent proportionally to the fraction of demand satisfied, with $r_{\text{DSR},t}(a_t) = 1$ indicating perfect demand satisfaction.

The second component, the overflow penalty term $r_{\text{of},t}(a_t)$, is designed to discourage unsafe operational states where tank levels approach their capacity limits. It is defined as:
\begin{equation}
    r_{\text{of},t}(a_t) = 
    \begin{cases}
      \sum_{l=1}^L \frac{h_{l,t} - \lambda_{\text{of}} \, h_{l,\text{max}}}{(1 - \lambda_{\text{of}})\, h_{l,\text{max}}} & \text{if } h_{l,t} > \lambda_{\text{of}} \, h_{l,\text{max}}, \\
      0 & \text{otherwise},
    \end{cases}
\end{equation}
where $h_{l,t}$ is the current level of tank $l$, $h_{l,\text{max}}$ is its maximum allowable level, and $\lambda_{\text{of}} \in (0,1)$ is a configurable threshold that defines the critical level for triggering the penalty. The penalty increases linearly beyond this threshold, encouraging the agent to avoid risky overfill conditions before they become critical.

\paragraph{Limitations.}
The main limitations of \texttt{WDSEnv} stem from the underlying simulator. First, from a modeling perspective, the environment does not allow for fine-grained control of pumps beyond simple on/off status, since features such as variable speed control are not supported by either Epynet or Epanet. 

Second, running experiments within this environment is computationally expensive, as reported in Table~\ref{tab:episode_time}. While our experiments are conducted on a relatively small network with a limited number of nodes and links, scaling to larger networks for benchmarking would require significantly more powerful hardware and the use of parallelization strategies. This challenge is inherent in the design of Epanet, which was not originally intended for real-time control or RL applications. Instead, Epanet was primarily developed to simulate the hydraulic behavior of water networks in response to predefined scenarios and configurations, such as in \textit{what-if} analyses.

Given these limitations, there is an opportunity for closer collaboration between the RL and hydraulic communities. A potential direction could involve the development of a lightweight, RL-friendly Python wrapper for Epanet, optimized for efficient, step-wise simulations and scalable training pipelines. Such a tool would significantly facilitate the integration of advanced learning algorithms into water system management tasks.

%% file: appendix/experiment_details.tex
\section{Experiment Details}
\subsection{\texttt{DamEnv}}
\label{sec:experiment_details:damenv}

In this section we provide more details on the benchmarking experiments performed on the \texttt{DamEnv} environment. In this setting we compare the SkRL~\cite{serrano2023skrl} version of PPO with some rule-based strategies described in Section~\ref{sec:dam}.
The parameters used for the PPO results reported in the main paper can be found in Table~\ref{tab:training_config_ppo_dam}. 
%To train PPO, it is possible to use the Jupyter notebook in \texttt{./examples/dam/benchmarks.ipynb} ore use the file in \texttt{./gym4real/algorithms/dam/ppo\_skrl.py}.

\begin{figure}[th!]
    \centering
    \includegraphics[width=0.7\linewidth]{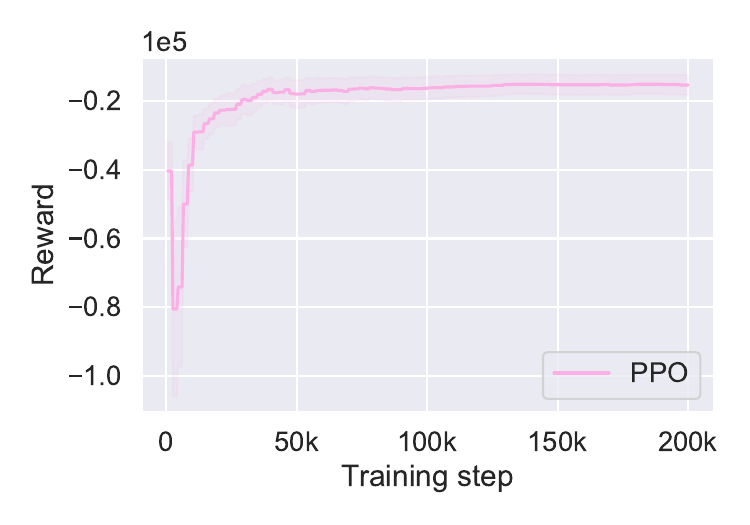}
    \caption{Learning curves of PPO on \texttt{DamEnv}.}
    \label{fig:dam_curves} 
\end{figure}

Additionally, Figure~\ref{fig:dam_curves} shows the validation performances during the training of PPO. In particular, the y-axis represents the mean return over $13$ year-long episodes, with $95$\%-confidence intervals. From the figure, we can observe the proper learning and convergence of the approach.

\begin{table}[ht]
    \centering
    \caption{Parameter configuration for PPO on \texttt{DamEnv}.}
    \vspace{0.1cm}
    \rowcolors{2}{gray!10}{white}
    \begin{tabular}{lll}
        \hline
        \rowcolor{gray!30}
        \textbf{Parameter} & \textbf{Value} \\
        \hline
        Training Steps & $200$k \\
        Batch Size & $32$ \\
       \# Epochs & $10$ \\
        Rollouts & $2048$ \\
        Gamma & $0.995$ \\
        Learning Rate & $8e-6$ \\
        Policy Network Size & $[16, 16]$ \\
        Initial log. std. & $-0.5$ \\
        Normalize obs. & True \\
        Seed & $123$ \\    
        \hline
    \end{tabular}
    \label{tab:training_config_ppo_dam}
\end{table}

\subsection{\texttt{ElevatorEnv}}
Within the \texttt{ElevatorEnv},  we compare custom implementation of tabular RL algorithms, i.e., Q-Learning and SARSA, against several rule-based strategies. Besides the trivial \textit{Random} policy, we wanted to provide also other two intuitive rule-based solutions that could be easily implemented in a physical system. Below, we briefly describe the rational behind such rule-based policies.
\begin{itemize}
    \item \textit{Random} policy: the agent selects an action uniformly at random, without considering the state of the system.
    \item \textit{Longest-first} (LF) policy: the agent prioritizes serving passengers on the floor with the longest waiting queue. Once passengers are picked up, the elevator travels directly to their destination, skipping any intermediate stops.
    \item \textit{Shortest-first} (SF) policy: the agent prioritizes serving passengers on the floor with the shortest waiting queue. Similarly, it proceeds directly to their destination floor after onboarding passengers without making additional stops.
\end{itemize}

As evidenced in Figures~\ref{fig:elevator-avg_reward} and~\ref{fig:elevator-boxplot}, both the LF and SF policies outperform the \textit{Random} one, demonstrating to be reasonable in this setting. However, Q-Learning and SARSA are able to achieve even higher performances both in their vanilla version and with a modest training. We expect even better performances from ad-hoc algorithms specifically designed to address this problem and its challenges.

\begin{figure}[ht]
    \centering
    \includegraphics[width=0.7\linewidth]{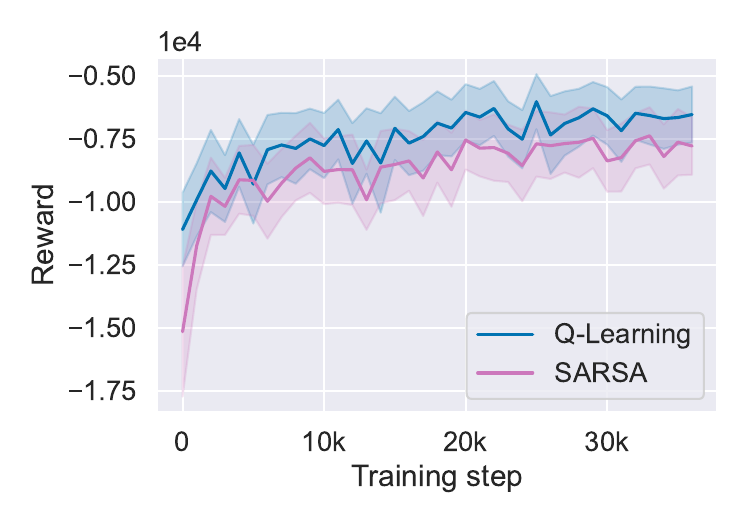} 
    \caption{Learning curves of Q-Learning and SARSA on \texttt{ElevatorEnv}.}
    \label{fig:elevator-curves} 
\end{figure}

We provide information on training configurations in Table~\ref{tab:elevator_alg_params} for both algorithms, as well as their learning curves in Figure~\ref{fig:elevator-curves}. In particular, y-axis reports the mean return, i.e., the \textit{global waiting time}, over $30$ episodes for each evaluation epoch, with a $95$\%-confidence interval. Notably, we trained Q-Learning and SARSA for $100,000$ episodes with an early-stopping mechanism that alts the training if after $10$ subsequent episodes no improvement grater than a threshold (\textit{tolerance}) occurs. 

\begin{table}[ht]
    \centering
    \caption{Parameter configuration for Q-Learning and SARSA on \texttt{ElevatorEnv}.}
    \vspace{0.1cm}
    \rowcolors{2}{gray!10}{white}
    \begin{tabular}{lll}
        \hline
        \rowcolor{gray!30}
        \textbf{Parameter} & \textbf{Q-Learning} & \textbf{SARSA} \\
        \hline
        \# Episodes & $100$k & $100$k \\
        \# Envs & 1 & 1 \\
        Gamma & $1.0$ & $1.0$ \\
        Epsilon & $1.0$ & $1.0$ \\
        Epsilon decay & $0.99$ & $0.99$ \\
        Minimum Epsilon & $0.05$ & $0.05$ \\
        Tolerance & $0.1$ & $0.1$ \\
        Stop after \# eps. with no improvement & $10$ & $10$ \\
        Seed & 42 & 42 \\ 
        \hline
    \end{tabular}
    \label{tab:elevator_alg_params}
\end{table}

\subsection{\texttt{MicrogridEnv}}
Within the \texttt{MicrogridEnv}, we compare the Stable-Baselines3~\cite{stable-baselines3} version of PPO against several rule-based strategies. Below, we briefly describe each of them.
\begin{itemize}
    \item \textit{Random} policy: the agent selects an action uniformly at random, without considering the current state of the system.
    \item \textit{Battery-first} (BF) policy: the agent prioritizes the use of the battery, attempting to store or retrieve energy as much as possible. The main grid is used only when the battery is insufficient ($a_t = 1$).
    \item \textit{Only-market} (OM) policy: the agent exclusively interacts with the main grid, buying and selling energy without utilizing the battery ($a_t = 0$). Note that the battery will still degrade over time due to temporal aging, even though it is not actively used.
    \item \textit{50-50} policy: the agent consistently splits the net power $P_{N,t}$ equally between the battery and the grid, by selecting $a_t = 0.5$ at every time step.
\end{itemize}

\begin{figure}[ht]
    \centering
    \includegraphics[width=0.7\linewidth]{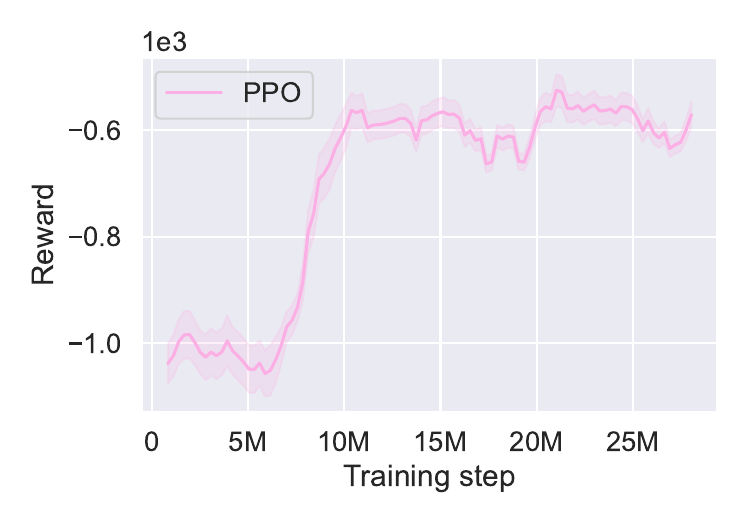} 
    \caption{Learning curves of PPO on \texttt{MicrogridEnv}.}
    \label{fig:mg-curves} 
\end{figure}

We trained PPO over $100$ episodes on $8$ parallel environments. The training span 4-year of data (from 2015 to 2019 included), while testing is done on year 2020, all with a resolution of $1$ hour.
Parameters of PPO used to obtained the results presented in the main paper are reported in Table~\ref{tab:mg_alg_params}. The learning curve of PPO is depicted in Figure~\ref{fig:mg-curves} with on y-axis the is reported the normalized mean episode return over 10 demand profiles, with $95$\%-confidence interval.

\begin{table}[ht]
    \centering
    \caption{Parameter configuration for PPO on \texttt{MicrogridEnv}.}
    \vspace{0.1cm}
    \rowcolors{2}{gray!10}{white}
    \begin{tabular}{ll}
        \hline
        \rowcolor{gray!30}
        \textbf{Parameter} & \textbf{Value} \\
        \hline
        \# Episodes & $100$ \\
        \# Envs & $8$ \\
        Policy Network Size & $[64,32]$ \\
        Gamma & $0.99$ \\
        Learning Rate & $5e{-5}$ \\
        Batch size & $512$ \\
        \# Epochs & $10$ \\
        Rollouts & $8912$ \\
        Initial log. std. & $-1$ \\
        Normalize obs. & True \\
        Seed & $42$ \\    
        \hline
    \end{tabular}
    \label{tab:mg_alg_params}
\end{table}

\subsection{\texttt{RoboFeederEnv}}\label{sec:experiment_details:robofeeder-picking}
This section gives further details on the training of RL agents on \texttt{RoboFeeder} for reproducibility purposes.
Table \ref{tab:experiment_details:robofeeder-ppo} enumerates the entire list of hyperparameters used to train the Stable-Baselines3~\cite{stable-baselines3} implementations of the PPO algorithm. This algorithm is used for both the \texttt{picking} and \texttt{planning} environments. % In order to train PPO agents, it is possible to use the provided scripts in the folder \texttt{./gym4real/algorithms/robofeeder/}.

\begin{table}[th] 
    \centering
    \caption{Parameters configuration for PPO on \texttt{RoboFeederEnv}.}
    \label{tab:experiment_details:robofeeder-ppo}
    \vspace{0.1cm}
    \rowcolors{2}{gray!10}{white}
    \begin{tabular}{lll}
        \hline
        \rowcolor{gray!30}
        \textbf{Parameter} & \textbf{PPO (\texttt{picking})} & \textbf{PPO (\texttt{planning})} \\
        \hline
        \# Episodes & $15$ & $10$ \\
        \# Envs & 20  & 6 \\
        Policy Network Size & $[512,128]$ & $[64,64]$ \\
        Gamma & $0.99$ & $0.99$ \\
        Learning Rate & $0.001$ & $0.0003$ \\
        Log Rate & $10$ & $10$ \\
        Batch Size & $128$ & $128$ \\
        \# Steps & $15$k & $10$k \\
        Ent. Coeff. & $0.01$ & $-$ \\
        Clip fraction & $0.2$ & $0.2$ \\
        Seed & [0,1,2,56,123] & [0,1,2,56,123] \\ 
        \hline
    \end{tabular}
    \label{tab:robofeeder_alg_params}
\end{table}

\subsubsection{\texttt{RoboFeeder-picking}} \label{sec:experiment_details:robofeederenv}

\begin{figure}[ht]
    \centering
    \begin{minipage}{0.49\linewidth}
    \subfloat[Mean Episode Length during training.]{
    \includegraphics[width=\linewidth]{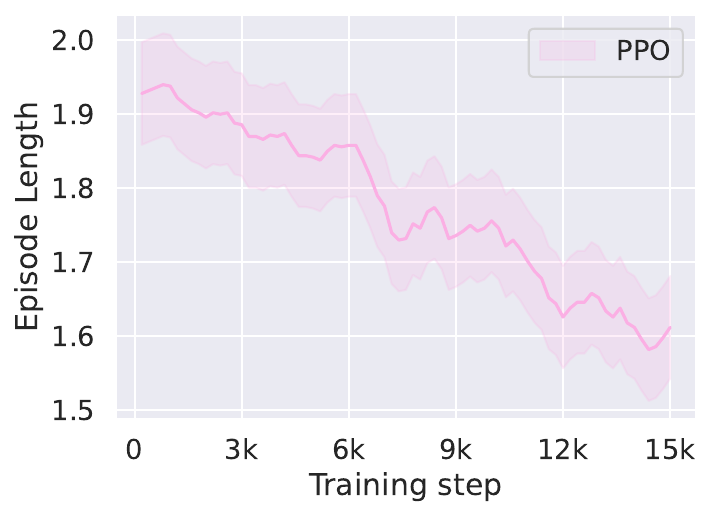}
    \label{fig:picking-ep_length}
    }
    \end{minipage}
    \hfill
     \begin{minipage}{0.49\linewidth}
     \subfloat[Mean reward during training.]{
     \includegraphics[width=\linewidth]{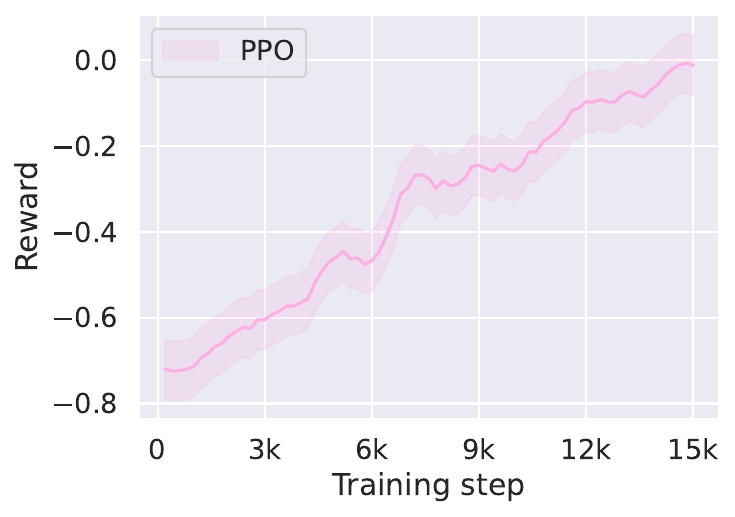}
     \label{fig:picking-curves}
     }
    \end{minipage}
    \caption{\texttt{RoboFeeder-picking-v0}, PPO training curve of mean episode length and reward over 5 seeds. 
    }
\end{figure}

Figures~\ref{fig:picking-ep_length} and~\ref{fig:picking-curves} illustrate the training curve of the PPO algorithm, interacting with the \texttt{RoboFeeder-picking-v0} environment. Based on $15$k timestamps, using a single object to train the robot to learn to pick the object, it is possible to notice how the distance-based reward helps the agent to learn the right approaching point to the object. The mean reward, over the multiple parallel environments considered, is normalized with respect to the total number of timesteps $T=2$, to have a value in the range $ r \in [-1,1]$. The training curves of the \texttt{robofeeder-picking-v1} are not provided, since they are in line with the ones provided for \texttt{v0}.
%construction of both the enviornment is based on the same logic.  
The training curve reported and the related confidence intervals are based on $5$ different seeds.

\subsubsection{\texttt{RoboFeeder-planning}} \label{sec:experiment_details:robofeeder-planning}
\begin{figure}[h]
    \centering
    \begin{minipage}{0.49\linewidth}
    \subfloat[Mean Episode Length during training.]{
    \includegraphics[width=\linewidth]{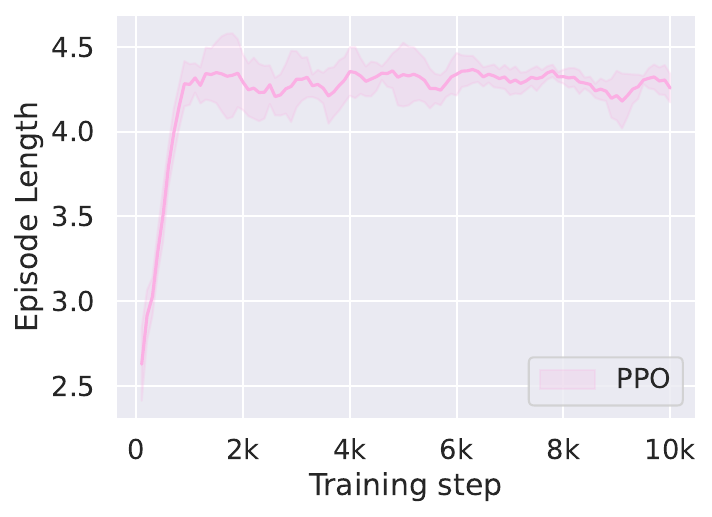}
    \label{fig:planning-ep_length}
    }
    \end{minipage}
    \hfill
     \begin{minipage}{0.49\linewidth}
     \subfloat[Mean reward during training.]{
     \includegraphics[width=\linewidth]{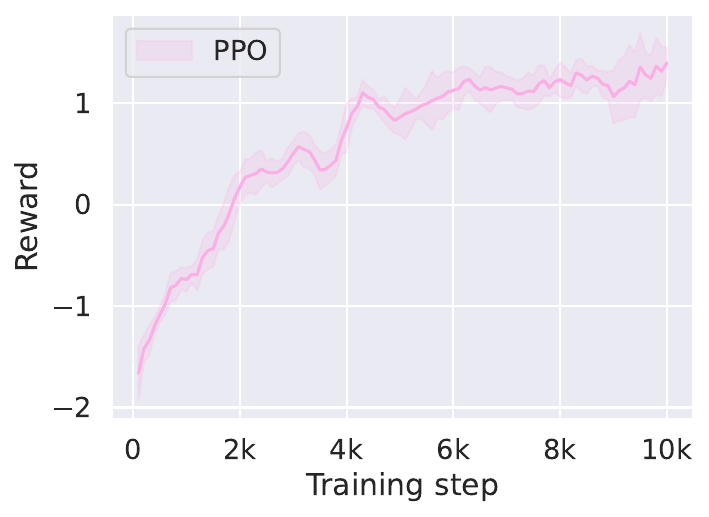}
     \label{fig:plannig-curves}
     }
    \end{minipage}
    \caption{\texttt{RoboFeeder-planning}, training curve of mean episode reward and episode length over 5 seeds.
    }
\end{figure}

Figures~\ref{fig:planning-ep_length} and~\ref{fig:plannig-curves} present the results of training the PPO algorithm over $10$k iterations. The environment is set up with three objects positioned so they can be successfully picked. Despite the limited number of training steps, the learning curve for the mean reward demonstrates the agent’s ability to select effective actions. With a total time horizon of $T=5$, the agent achieves positive rewards, indicating successful object picking. Initially, episodes end prematurely, as shown in the mean episode length plot. However, over time, the agent converges to an optimal time horizon of $T=4$, which aligns with picking the three available objects, followed by a reset action once the scene is empty. The resulting training curves are based on $5$ different seeds.

\subsection{\texttt{TradingEnv}}
\label{sec:experiment_details:tradingenv}

This section gives further details on the training of RL agents on \texttt{TradingEnv} for reproducibility purposes.
%First of all, it is necessary to download EUR/USD data from the website: \url{https://www.histdata.com/download-free-forex-historical-data/?/ascii/tick-data-quotes/EURUSD}, from 2019 to 2022. All the downloaded files must then be inserted in the folder \texttt{./gym4real/data/trading/forex}.
Regarding the environment configuration, we considered the base setting, as referred in the Appendix~\ref{sec:experiment_details:tradingenv}, using 60 delta mid-prices, the timestamp, and the position of the agent. The trading activity is allowed from 8:00 EST to 18:00 EST. Due to the use of Deep RL algorithms as PPO~\citep{schulman2017ppo} and DQN~\citep{Mnih2015Atari}, the timestamp is transformed using cyclic encoding. 

\begin{figure}[h]
    \centering
    \begin{minipage}{0.49\linewidth}
    \subfloat[Episode Reward Mean during PPO Training.]{
    \includegraphics[width=\linewidth]{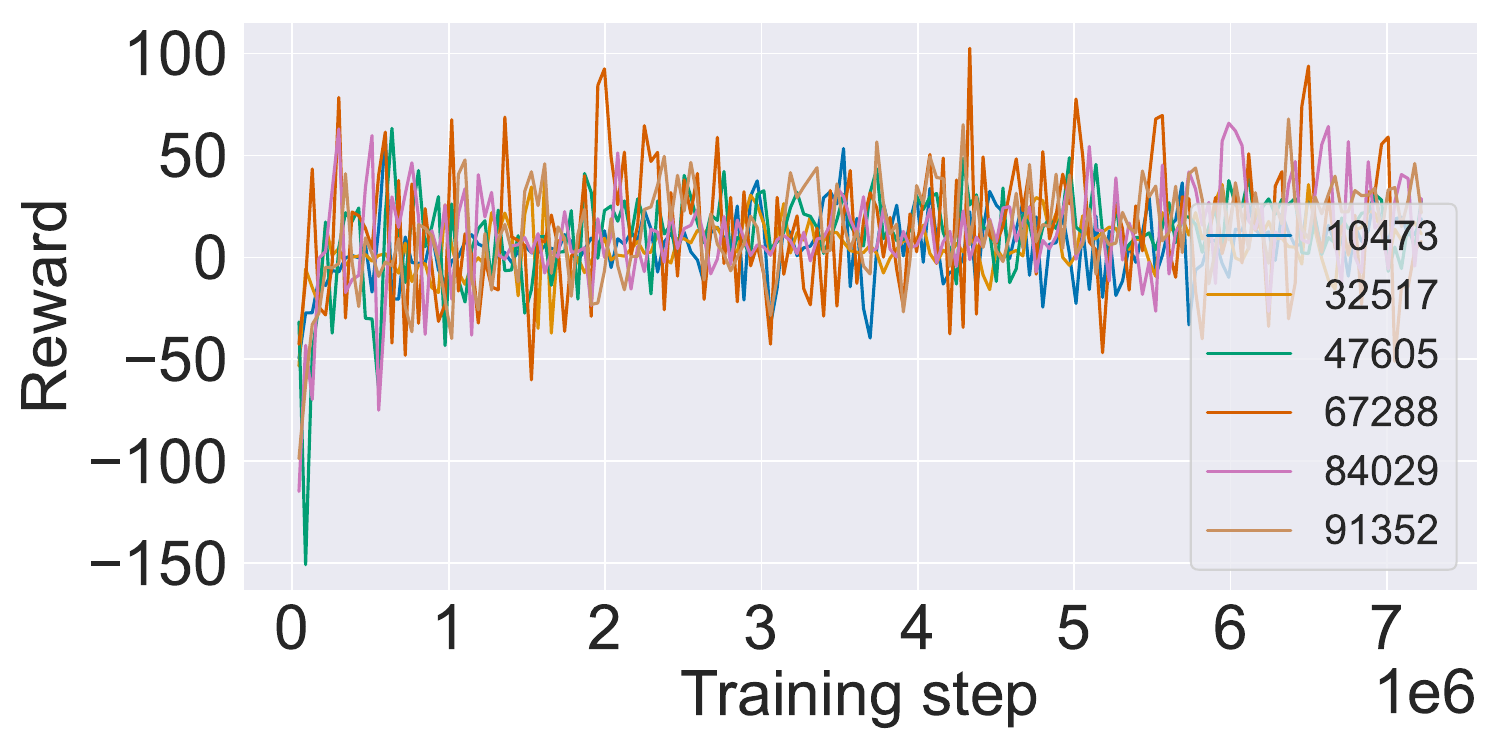}
    \label{fig:training_curves_ppo-appendix:train}
    }
    \end{minipage}
    \hfill
     \begin{minipage}{0.49\linewidth}
     \subfloat[Cum. Reward of PPO on validation data.]{
     \includegraphics[width=\linewidth]{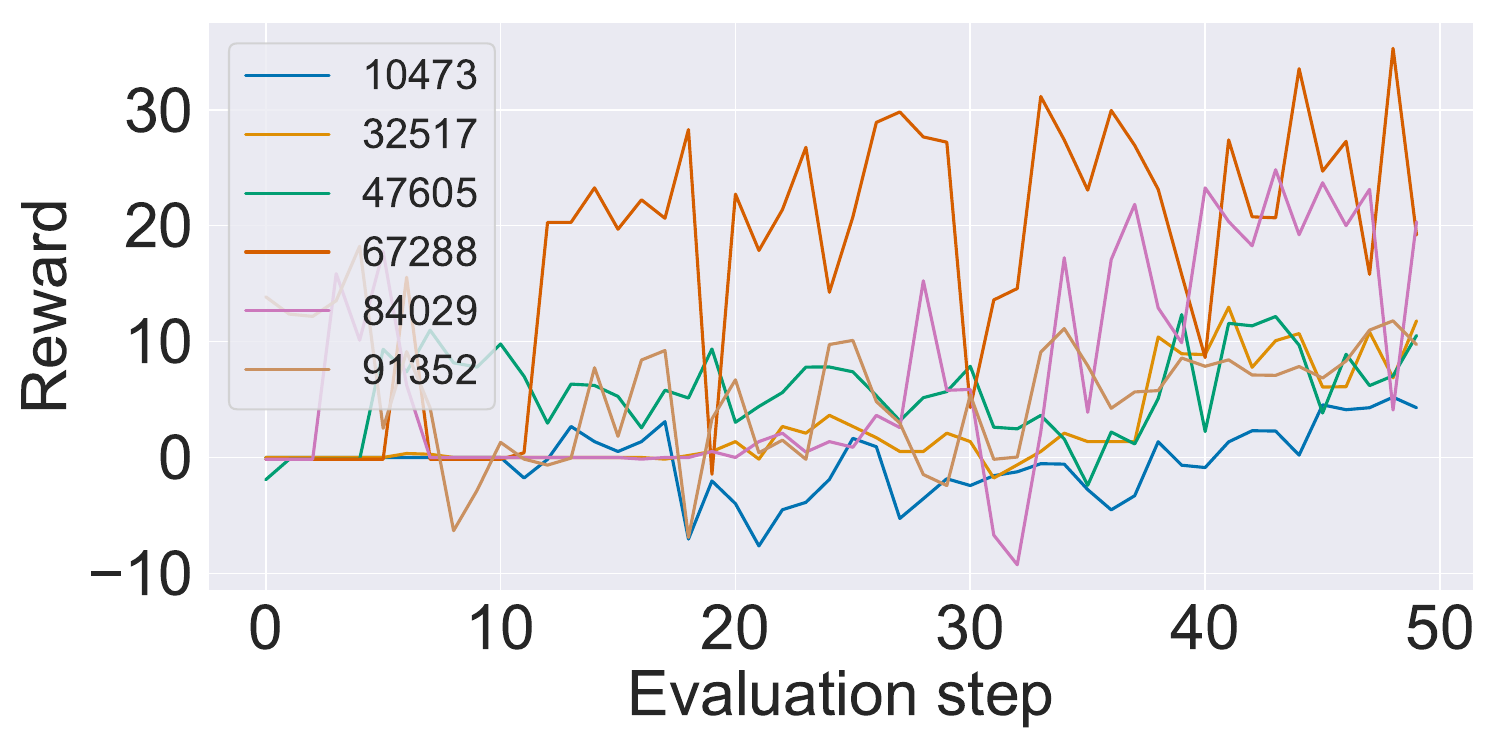}
     \label{fig:training_curves_ppo-appendix:eval}
     }
    \end{minipage}
    \caption{Training Curves PPO of the 6 seeds considered for \texttt{TradingEnv}, (a) episode reward mean on Training and (b) cumulative reward on the entire validation year.
    }
    \label{fig:training_curves_ppo-appendix}
\end{figure}

\begin{figure}[h]
    \centering
    \begin{minipage}{0.49\linewidth}
     \subfloat[Episode Reward Mean during DQN Training.]{
        \includegraphics[width=\linewidth]{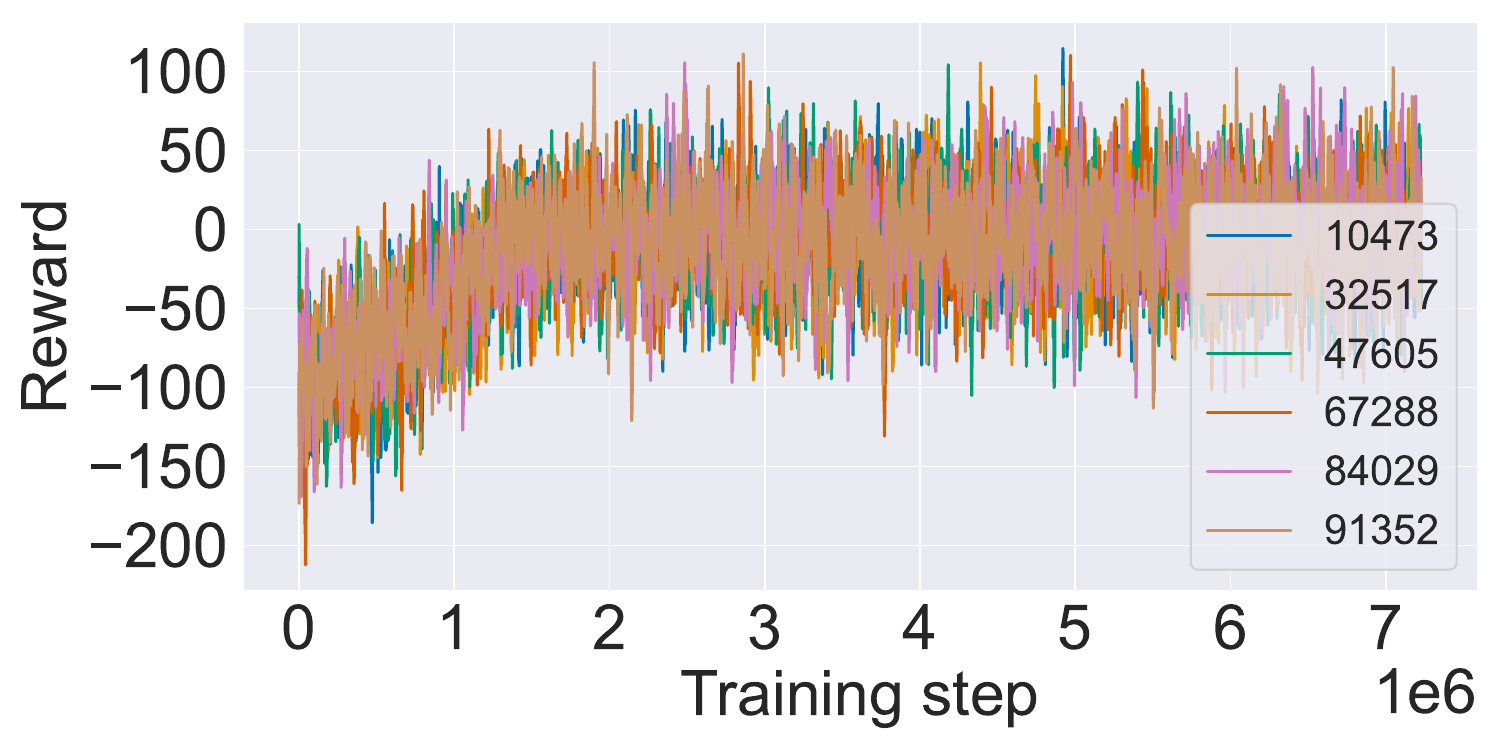}
        }
        \label{fig:training_curves_dqn-appendix:train}
    \end{minipage}
    \hfill
     \begin{minipage}{0.49\linewidth}
      \subfloat[Cum. Reward of DQN on Validation.]{
        \includegraphics[width=\linewidth]{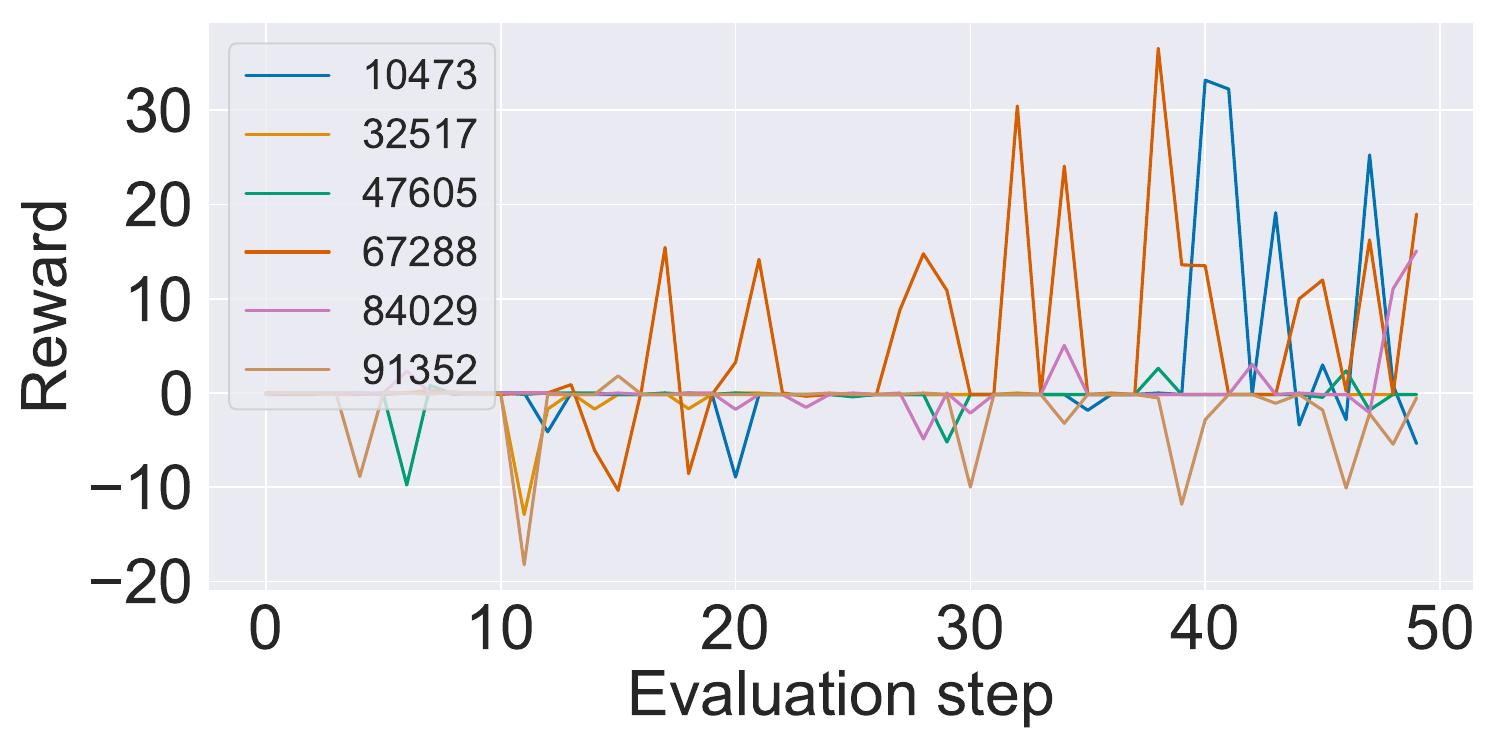}
        }
        \label{fig:training_curves_dqn-appendix:eval}
    \end{minipage}
    \caption{Training Curves DQN of the 6 seeds considered for \texttt{TradingEnv}, (a) episode reward mean on Training and (b) cumulative reward on the entire validation year.
    }
    \label{fig:training_curves_dqn-appendix}
\end{figure}

%The exact environment configuration used in the experiments can be found in the files \texttt{./gym4real/envs/trading/world\_train}, \texttt{./gym4real/envs/trading/world\_validation} and \texttt{./gym4real/envs/trading/world\_test}. 
It is important to notice that training, validation, and tests were performed using different sets of years on 6 seeds. In particular, the model was trained from 2019 to 2020, validated in 2021, where the best model is selected, and tested in 2022.
The hyperparameter configurations chosen for DQN and PPO are reported in Table~\ref{tab:trading_alg_params}, while the related training curves can be found in Figure~\ref{fig:training_curves_ppo-appendix} and~\ref{fig:training_curves_dqn-appendix}.
We used the Stable-Baselines3~\cite{stable-baselines3} implementations of the algorithms, specifically the SBX (Stable-Baselines3 in JAX) implementation of PPO and the standard SB3 implementation of DQN.

%\begin{verbatim}
%    args = {
%        'exp_name': 'trading/ppo',
%        'n_episodes': 30,
%        'n_envs': 6,
%        'policy_kwargs': dict(net_arch=[512, 512]),
%        'gamma': 0.90,
%        'learning_rate': 0.0001,
%        'log_rate': 10,
%        'batch_size': 236,
%        'n_steps':  118*6,
%        'ent_coeff': 0.,
%        'seeds': [32517, 84029, 10473, 67288, 91352, 47605] 
%    }
%\end{verbatim}

\begin{table}[h!]
    \centering
    \caption{Parameter configuration for PPO and DQN on \texttt{TradingEnv}.}
    \vspace{0.1cm}
    \rowcolors{2}{gray!10}{white}
    \begin{tabular}{lll}
        \rowcolor{gray!30}
        \hline
        \textbf{Parameter} & \textbf{PPO} & \textbf{DQN} \\
        \hline
        \# Episodes & $30$ & $30$ \\
        \# Envs & 6 & 6 \\
        Policy Network Size & $[512, 512]$ & $[512, 512]$ \\
        Gamma & $0.90$ & $0.90$ \\
        Learning Rate & $0.0001$ & $0.0001$ \\
        Log Rate & $10$ & - \\
        Batch Size & $236$ & $64$ \\
        \# Steps & $708$ & - \\
        Entropy Coeff. & $0.0$ & - \\
        Buffer Size & - & $1$M \\
        Learning Starts & - & $100$ \\
        Exploration Fraction & - & $0.2$ \\
        Exploration Final Eps. & - & $0.05$ \\
        Polyak Update (tau) & - & 1.0 \\
        Train Frequency & - & $4$ \\
        \hline
    \end{tabular}
    \label{tab:trading_alg_params}
\end{table}

Standard hyperparameters have been used for benchmarking, hence we encourage the use of hyperparameter tuning techniques to better calibrate the model. Indeed, without a proper methodology for hyperparameter tuning and model selection, it is very easy to overfit when manually selecting their values.

%To train or evaluate PPO and DQN, it is possible to use the notebook available in \texttt{./examples/trading/benchmark.ipynb}. In particular, it is possible to retrain the models or load pre-trained models by setting the appropriate flag and see the performances on training, validation and test years.

%In order to train PPO and DQN agents, it is also possible to use the scripts provided in the folder \texttt{./gym4real/algorithms/trading/}.

\begin{figure}[h]
    \centering
    \begin{minipage}{0.49\linewidth}
    \subfloat[Training Years.]{
    \includegraphics[width=\linewidth]{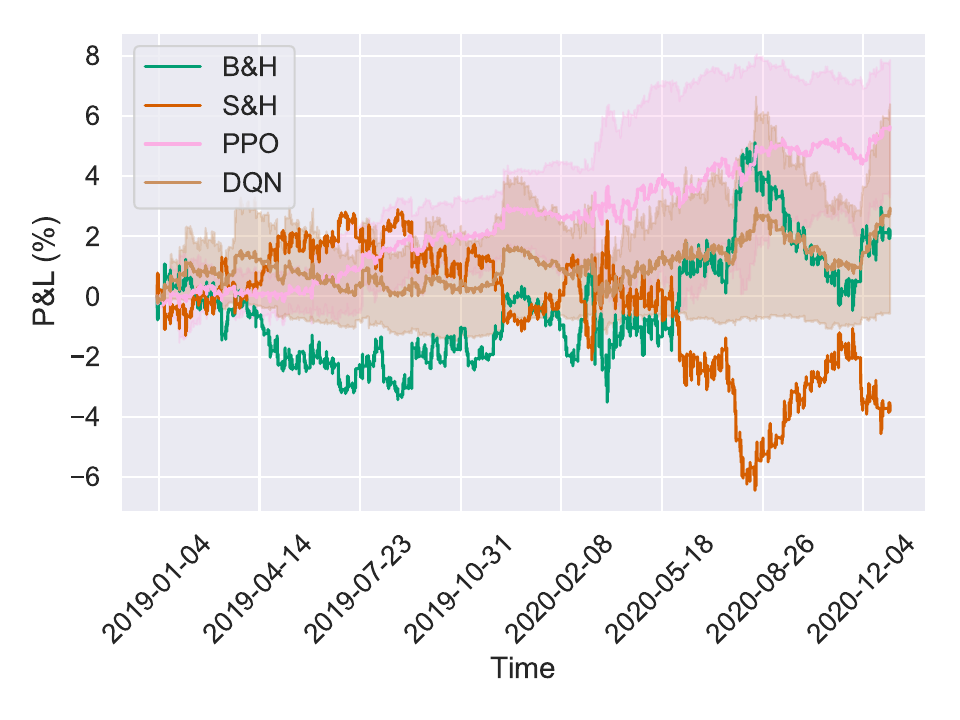}
    }
    \end{minipage}
    \hfill
    \begin{minipage}{0.49\linewidth}
    \subfloat[Validation Years.]{
        \includegraphics[width=\linewidth]{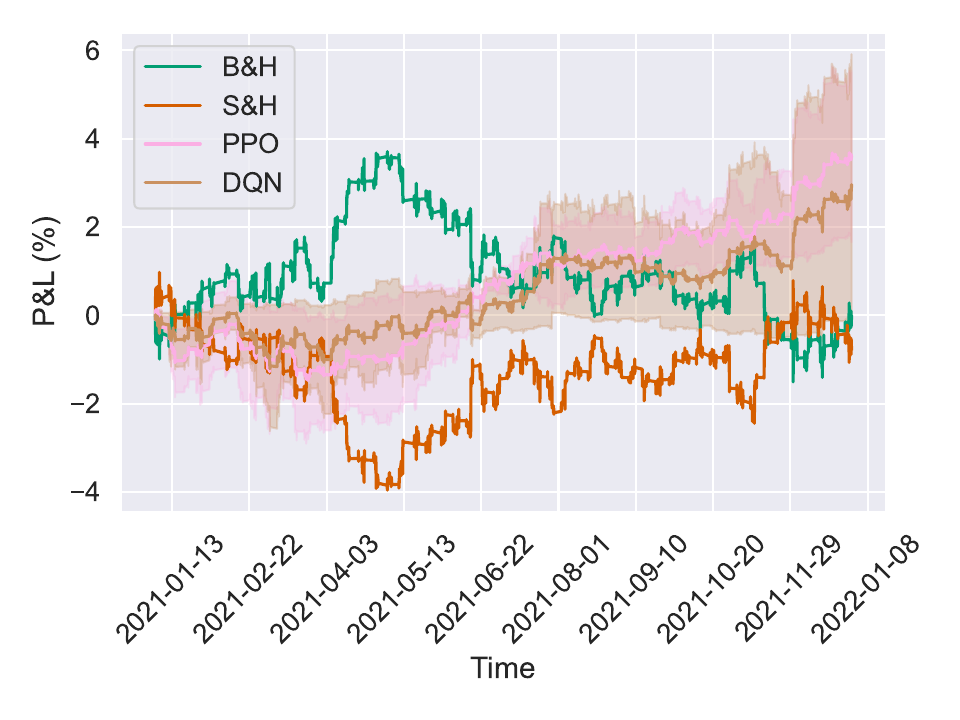}
        }
    \end{minipage}
    \caption{Performances of PPO and DQN w.r.t. common baselines (i.e., passive strategies) on Training (2019-2020) and Validation (2021) for \texttt{TradingEnv}. Confidence intervals obtained with 6 seeds.}
    \label{fig:tradingenv_PPODQNvsBaseline-appendix}
\end{figure}

In Figure~\ref{fig:tradingenv_PPODQNvsBaseline-appendix}, the P\&L performances for Training, Validation, and Test years are reported in addition to common passive strategies. The training was performed for approximately 7M steps. Unlike other domains, these baseline strategies are particularly strong. Indeed, people who invest in the stock market typically adopt the Buy\&Hold strategy, buying the stock and keeping the position long for years.
In this environment, the Buy\&Hold policy corresponds to consistently selecting the \textit{Long} action, while the Sell\&Hold policy corresponds to consistently selecting the \textit{Short} action.

\begin{figure}[h]
    \centering
    \begin{minipage}{0.32\linewidth}
        \includegraphics[width=\linewidth]{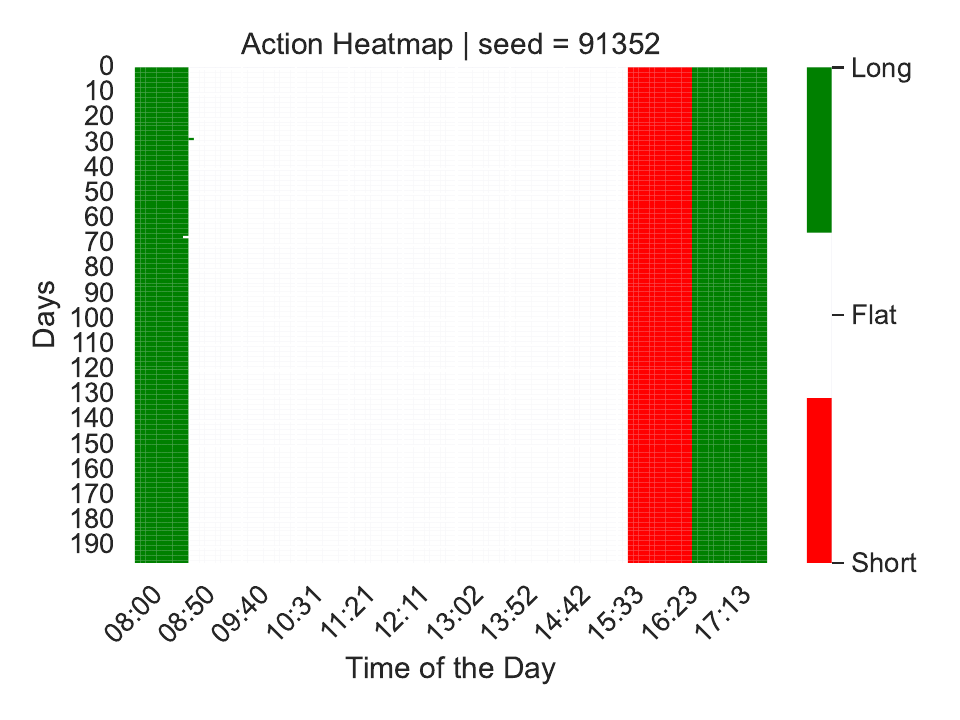}
    \end{minipage}
    \begin{minipage}{0.32\linewidth}
        \includegraphics[width=\linewidth]{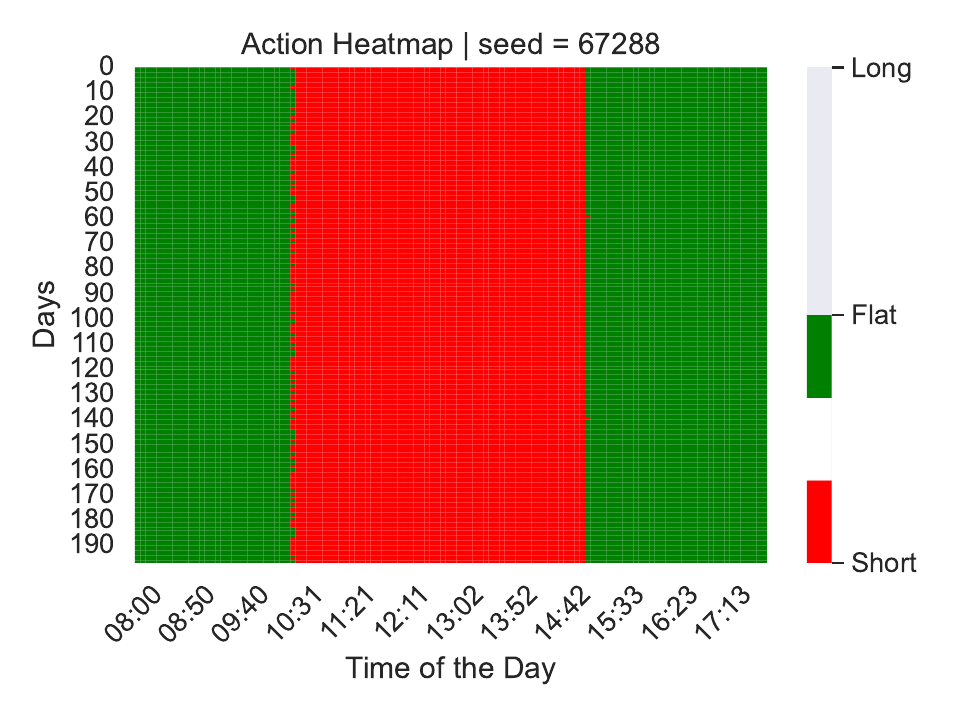}
    \end{minipage}
     \begin{minipage}{0.32\linewidth}
        \includegraphics[width=\linewidth]{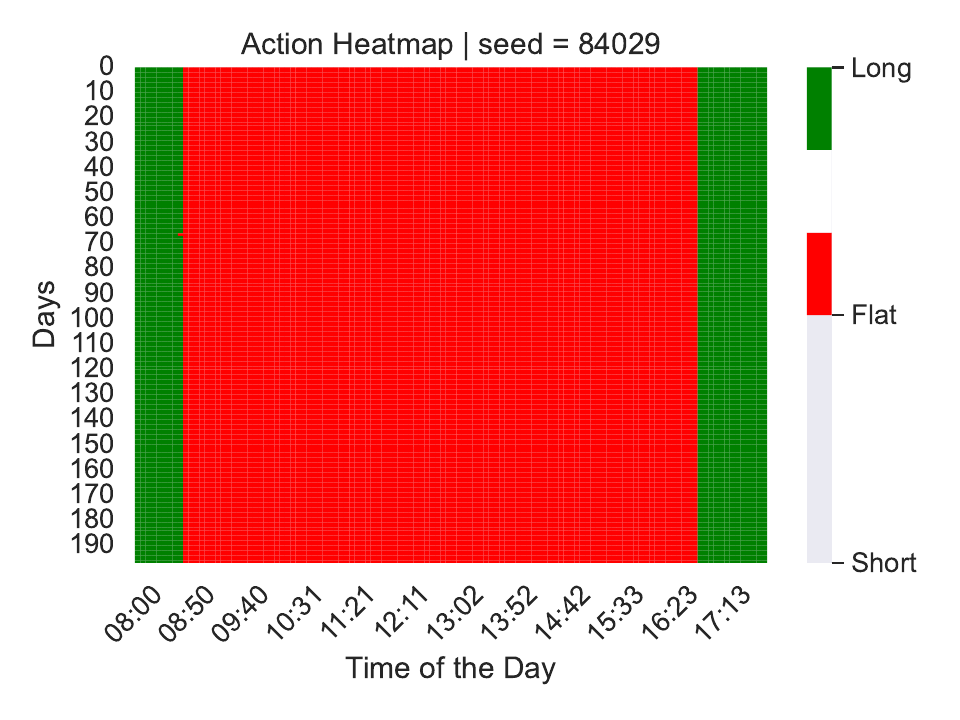}
    \end{minipage}
    \caption{Policy learned by PPO on 3 different seeds in the \texttt{TradingEnv}. The x-axis and the y-axis report the time of the day and the different days, respectively. Noticeably, three different policies are learned depending on the seed.}
    \label{fig:tradingenv_learned_policy-appendix}
\end{figure}

Figure~\ref{fig:tradingenv_learned_policy-appendix} shows a sample of three policies on the validation set with PPO. We can observe that three different seeds lead to three significantly different policies, underlying the high stochasticity of the environment.

\subsection{\texttt{WDSEnv}}
Within the \texttt{WDSEnv}, we compare the Stable-Baselines3~\cite{stable-baselines3} version of DQN against different rule-based strategies. Hereafter, we spend a few words to describe each.
\begin{itemize}
    \item \textit{Random} policy: the agent selects an action uniformly at random, without considering the state of the system;
    \item \textit{P78} policy: the agent plays a fixed action $a_t = 2$, keeping the pump P78 always open and the pump P79 always closed;
    \item \textit{P79} policy: conversely from the previous policy, the agent plays a fixed action $a_t = 1$, keeping the pump P79 always open and the pump P78 always closed;
    \item \textit{Default} policy: the agent plays a strategy defined in the \textit{[CONTROLS]} section of the \textit{.inp} file. This represents an expert policy adopted by default within Epanet. In this case, the strategy requires the following actions: 
    \begin{itemize}
        \item with $l$ being tank \textit{T41}, open P78 if $h_{l,t} < 5.0$ and close P78 if $h_{l,t} > 8.0$;
        \item with $l$ being tank \textit{T42}, open P79 if $h_{l,t} < 5.0$ and close P79 if $h_{l,t} > 8.0$.
    \end{itemize}
\end{itemize}

In this environment, we adopt the DQN algorithm, which is well-suited to the setting due to the continuity of the observation space and the finite nature of the action space. Training is carried out over episodes of one-week duration, with a control time step of one hour. Each episode presents the agent with varying demand profiles, sampled according to the occurrence probabilities reported in Table~\ref{tab:wds_params}. Experimental results show that DQN consistently outperforms all rule-based strategies, as it learns to execute more conservative control actions that effectively reduce the risk of tank overflow, while still maintaining a high DSR.

\begin{figure}[ht]
    \centering
    \includegraphics[width=0.7\linewidth]{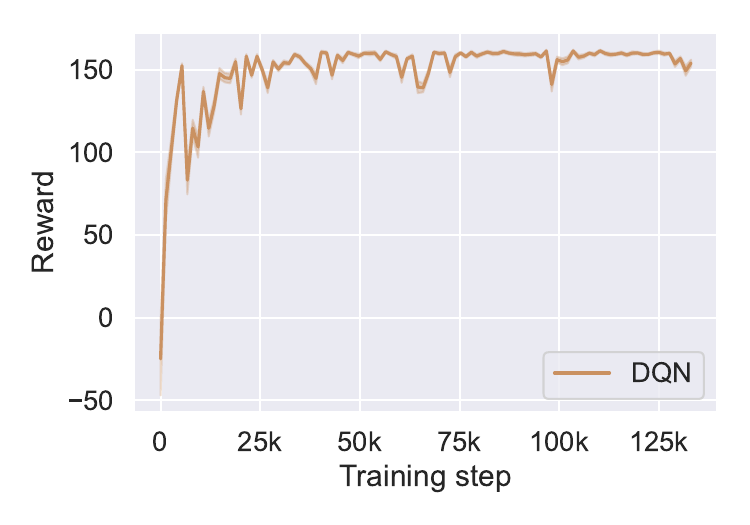} 
    \caption{Learning curves of DQN on \texttt{WDSEnv}.}
    \label{fig:wds-curves} 
\end{figure}

The full training configuration for DQN is detailed in Table~\ref{tab:wds_alg_params}, while the convergence of the algorithm is illustrated in Figure~\ref{fig:wds-curves}. The y-axis in the figure reports the mean episode return over 10 evaluation episodes, with a 95\% confidence interval.

\begin{table}[ht]
    \centering
    \caption{Parameter configuration for DQN on \texttt{WDSEnv}.}
    \vspace{0.1cm}
    \rowcolors{2}{gray!10}{white}
    \begin{tabular}{ll}
        \hline
        \rowcolor{gray!30}
        \textbf{Parameter} & \textbf{Value} \\
        \hline
        \# Episodes & $100$ \\
        \# Envs & $8$ \\
        Policy Network Size & $[64,64]$ \\
        Gamma & $0.99$ \\
        Learning Rate & $0.001$ \\
        Batch Size & $32$ \\
        Buffer Size & $1$M \\
        Learning Starts & 100 \\
        Exploration Fraction & $0.1$ \\
        Exploration Final Eps. & $0.05$ \\
        Polyak update (tau) & 1.0 \\
        Train Frequency & $4$ \\
        Seed & $42$ \\    
        \hline
    \end{tabular}
    \label{tab:wds_alg_params}
\end{table}

